\documentclass[twoside,journey]{IEEEtran}
\usepackage{makecell}
\usepackage{hyperref}
\usepackage{array}
\usepackage{graphicx,amssymb,amsmath}
\usepackage{multicol}
\usepackage[noadjust]{cite}
\usepackage{setspace}
\usepackage{subcaption}
\usepackage{graphicx}
\usepackage{float}
\usepackage {url}
\usepackage{stfloats}
\usepackage{amsthm,pifont}
\usepackage{flushend}
\usepackage{cases,subeqnarray}
\usepackage{bm,multirow,bigstrut}
\usepackage{amsmath, amsthm, amssymb}
\usepackage{textcomp}
\usepackage{latexsym,bm}
\usepackage{booktabs}
\usepackage{xcolor}
\usepackage{booktabs} 
\usepackage{mathtools}
\usepackage{dsfont}
\usepackage{extarrows}
\usepackage{epsfig}
\usepackage{epsfig}
\usepackage{epstopdf}
\usepackage[noend]{algpseudocode}
\usepackage{algorithmicx,algorithm}
\usepackage{makecell}
\usepackage{colortbl}
\usepackage{booktabs}
\usepackage{graphicx} 
\usepackage{array}    
\usepackage{hyperref} 
\usepackage{booktabs}
\usepackage{enumitem}
\usepackage{diagbox}
\usepackage{tikz}
\usetikzlibrary{trees,shadows.blur}
\usepackage{forest}

\theoremstyle{plain}

\theoremstyle{plain}

\usepackage{amsmath}

\newcommand{\ignore}[1]{{{\color{yellow} }}}
\definecolor{blue-green}{rgb}{0.0, 0.87, 0.87}
\IEEEoverridecommandlockouts
\begin{document}

\title{SecDiff: Diffusion-Aided Secure Deep Joint Source-Channel Coding Against Adversarial Attacks}
\author{Changyuan Zhao, Jiacheng Wang, Ruichen Zhang, Dusit Niyato,~\IEEEmembership{Fellow,~IEEE}, Hongyang Du,\\
Zehui Xiong,~\IEEEmembership{Senior Member,~IEEE}, Dong In Kim,~\IEEEmembership{Life Fellow,~IEEE}, Ping Zhang,~\IEEEmembership{Fellow,~IEEE}
\thanks{C. Zhao is with the College of Computing and Data Science, Nanyang Technological University, Singapore, and CNRS@CREATE, 1 Create Way, 08-01 Create Tower, Singapore 138602 (e-mail: zhao0441@e.ntu.edu.sg).
}
\thanks{J. Wang, R. Zhang, and D. Niyato are with the College of Computing and Data Science, Nanyang Technological University, Singapore (e-mail: jiacheng.wang@ntu.edu.sg, ruichen.zhang@ntu.edu.sg, dniyato@ntu.edu.sg).}
\thanks{H. Du is with the Department of Electrical and Electronic Engineering, University of Hong Kong, Pok Fu Lam, Hong Kong SAR, China
(e-mail: duhy@eee.hku.hk).}
\thanks{Z. Xiong is with the School of Electronics, Electrical Engineering and Computer Science, Queen’s University Belfast, Belfast, BT7 1NN,
U.K. (z.xiong@qub.ac.uk).}
\thanks{D. I. Kim is with the Department of Electrical and Computer
Engineering, Sungkyunkwan University, Suwon 16419, South Korea (e-mail: dongin@skku.edu).}
\thanks{P. Zhang is with the State Key Laboratory of
Networking and Switching Technology, Beijing University of Posts and
Telecommunications, China (e-mail: pzhang@bupt.edu.cn).}
}

\maketitle
\vspace{-1cm}

\begin{abstract}
Deep joint source-channel coding (JSCC) has emerged as a promising paradigm for semantic communication, delivering significant performance gains over conventional separate coding schemes. However, existing JSCC frameworks remain vulnerable to physical-layer adversarial threats, such as pilot spoofing and subcarrier jamming, compromising semantic fidelity.
In this paper, we propose SecDiff, a plug-and-play, diffusion-aided decoding framework that significantly enhances the security and robustness of deep JSCC under adversarial wireless environments. Different from prior diffusion-guided JSCC methods that suffer from high inference latency, SecDiff employs pseudoinverse-guided sampling and adaptive guidance weighting, enabling flexible step-size control and efficient semantic reconstruction.
To counter jamming attacks, we introduce a power-based subcarrier masking strategy and recast recovery as a masked inpainting problem, solved via diffusion guidance. For pilot spoofing, we formulate channel estimation as a blind inverse problem and develop an expectation-minimization (EM)-driven reconstruction algorithm, guided jointly by reconstruction loss and a channel operator. Notably, our method alternates between pilot recovery and channel estimation, enabling joint refinement of both variables throughout the diffusion process.
Extensive experiments over orthogonal frequency-division multiplexing (OFDM) channels under adversarial conditions show that SecDiff outperforms existing secure and generative JSCC baselines by achieving a favorable trade-off between reconstruction quality and computational cost. This balance makes SecDiff a promising step toward practical, low-latency, and attack-resilient semantic communications.

\end{abstract}
\IEEEpeerreviewmaketitle

\section{Introduction}\label{intro}

















In recent years, deep learning has demonstrated remarkable success in wireless communications, delivering substantial performance improvements over traditional analytical methods~\cite{zhao2024generative, liu2024semantic, wang2025generative}. Among these advances,
deep joint source-channel coding (JSCC) has emerged as a transformative paradigm for end-to-end learned communication. In deep JSCC, neural networks replace the conventional separate modules for source compression and channel coding, enabling a unified encoder-decoder architecture trained directly over the communication channel~\cite{bourtsoulatze2019deep, xu2021wireless, erdemir2023generative}. By directly mapping input data, such as image or sensor data, to channel symbols and optimizing the entire pipeline jointly, deep JSCC adapts transmission strategies to both the content and channel characteristics. This approach yields significant advantages in the finite-blocklength regime, outperforming traditional separate coding schemes under realistic channel conditions~\cite{gunduz2024joint}. As a result, deep JSCC has been successfully extended to various wireless scenarios, including image transmission, multi-antenna systems, and multi-user networks, consistently demonstrating improved robustness and efficiency over classical designs~\cite{wu2024deep}. Furthermore, it aligns with the emerging paradigm of semantic communications, where the goal shifts from accurate bit-level recovery to preserving the semantic fidelity of transmitted content~\cite{nguyen2025contemporary, huang2025d, lyu2024semantic}.

Recently, researchers have begun leveraging generative AI (GenAI), especially diffusion models, to further enhance deep JSCC performance~\cite{zhang2025semantics,wu2024cddm, wang2025diffcom}. Diffusion models are powerful generative models that capture the data distribution and can iteratively denoise noisy inputs into high-quality outputs~\cite{yang2023diffusion}. By tapping into this generative capability as a prior, diffusion-enhanced JSCC schemes enable more semantic-aware recovery under lossy or degraded channels. For example, some works treat the distortion in deep JSCC outputs as an image restoration problem: a coarse reconstruction is first obtained via Deep JSCC, then a pre-trained diffusion model refines it to restore fine details~\cite{yilmaz2024high}. Such methods have achieved remarkable fidelity improvements, even in extremely poor channel conditions where conventional decoders would fail~\cite{zhang2025semantics}. 
Overall, these generative model-guided JSCC approaches illustrate the potential of combining semantic-level knowledge with physical-layer encoding to push performance beyond what purely discriminative models can achieve~\cite{van2024generative}.


Despite their promise, existing JSCC methods face several security issues:
\begin{enumerate}
    \item \textbf{Susceptibility to pilot spoofing:}
Many JSCC methods critically depend on accurate Channel State Information (CSI) to guide encoding strategies~\cite{yang2024swinjscc, kurka2020deepjscc}. This design assumption makes them particularly vulnerable to pilot spoofing attacks, where an adversary transmits forged pilot signals to manipulate the estimated CSI at the receiver~\cite{tugnait2018pilot}. Once the CSI is misled, the generative decoder operates under false assumptions about the channel, leading to severe performance degradation. 
    \item  \textbf{Fragility to jamming attacks:}
Most existing deep JSCC models are optimized under assumptions of benign channel noise and do not account for active jamming~\cite{bourtsoulatze2019deep}. However, an intelligent attacker can disrupt communication by jamming critical subcarriers, corrupting the transmitted signal in a way that severely distorts the latent representations processed by the decoder~\cite{nan2023physical}. 
    \item \textbf{Intractability of secure real-time decoding:}
GenAI-aided, especially diffusion-based, JSCC methods require multiple iterative steps to reconstruct each sample, resulting in substantial latency and computational overhead~\cite{wu2024cddm}.
In adversarial environments where pilot spoofing or jamming may dynamically alter the effective channel state, the decoder may require additional correction or resampling, compounding the delay~\cite{wang2025diffcom}. 
\end{enumerate}

In this work, we consider a wireless semantic communication scenario with deep JSCC operating over orthogonal frequency-division multiplexing (OFDM)-based multipath fading channels. We aim to address the challenge of maintaining high-fidelity semantic decoding in the presence of two physical-layer adversarial threats: pilot spoofing (global contamination) and subcarrier jamming (partial corruption). 
Our goal is to design a general and low-latency decoding strategy that can resist attacks and enhance semantic fidelity.
Therefore, we propose SecDiff, a plug-and-play,
diffusion-aided JSCC decoding architecture that overcomes these challenges. Our approach is plug-and-play and computationally efficient, integrating a diffusion model into the JSCC decoder via pseudoinverse-guided sampling and step-skipping sampling.
Pseudoinverse-guided sampling leverages channel structure to better recover information from jammed subcarriers, enhancing robustness against adversarial interference.
Step-skipping sampling accelerates the diffusion process by reducing the number of denoising steps, thereby achieving high-fidelity reconstruction with significantly lower latency. Distinct from prior works that rely on clean CSI or lack adversarial robustness, SecDiff is explicitly designed to defend against both subcarrier jamming and pilot spoofing. To this end, we introduce a power-based subcarrier masking strategy and formulate recovery as a masked inpainting problem, while also developing an expectation-minimization (EM)-driven channel estimation framework to support secure decoding under spoofed pilot contamination.
\begin{itemize}
    \item We propose \textit{SecDiff}, a plug-and-play diffusion-based decoding framework that integrates pseudoinverse-guided sampling and adaptive step-size control into the JSCC pipeline. By leveraging step-skipping sampling, SecDiff achieves high-fidelity reconstruction quality with significantly reduced inference latency and computational overhead.
    \item To defend against partial-band jamming, we introduce a power-based subcarrier masking mechanism and formulate the decoding task as a masked inpainting problem. The proposed masked diffusion guidance effectively reconstructs semantic information over jammed subcarriers using the spectral context of clean ones.

   \item We reformulate the joint signal and channel recovery under pilot spoofing as a blind inverse problem and propose an EM-driven diffusion decoding strategy. The algorithm alternates between estimating the clean signal and refining the channel operator, minimizing reconstruction error while enforcing regularization. This iterative refinement enables robust semantic decoding without relying on explicit CSI.

    \item 
    SecDiff achieves a superior quality–latency trade-off by improving PSNR by 4.4\% and reducing perceptual distortion by over 30\% compared to the fastest diffusion baseline under jamming. Under pilot spoofing, SecDiff doubles the success rate of benchmark methods and boosts PSNR by 19.3\% with 36.1\% lower FID, establishing itself as a practical, low-latency, and attack-resilient semantic communication framework.
    
\end{itemize}

\section{Related Work}

\subsection{Deep JSCC for Wireless Communications}

JSCC with deep learning has garnered significant attention since its inception. The seminal DeepJSCC framework proposed by Bourtsoulatze et al.~\cite{bourtsoulatze2019deep} demonstrated that an end-to-end autoencoder could outperform traditional separate source and channel coding schemes for image transmission over both additive white Gaussian noise (AWGN) and Rayleigh fading channels. Building upon this foundation, numerous neural architectures have been investigated to enhance the efficiency and adaptability of JSCC, including Vision Transformers~\cite{wu2024transformer}, Swin Transformers~\cite{yang2024swinjscc}, and nonlinear transform coding methods~\cite{dai2022nonlinear}.
In parallel, deep JSCC has been extended to more challenging communication scenarios such as MIMO fading channels~\cite{wu2024deep}, OFDM systems~\cite{yang2021deep, wu2022channel}, multi-hop relay networks~\cite{bian2025process}, and multi-user environments~\cite{yilmaz2023distributed}. Across these diverse settings, deep JSCC consistently achieves lower end-to-end distortion for a given bandwidth and power, particularly in low-SNR or short-packet regimes where conventional coding techniques often underperform~\cite{zhang2025semantics}.
Moreover, several recent designs have demonstrated the adaptability of deep JSCC models, enabling a single trained model to dynamically adjust to varying channel conditions or target compression rates~\cite{dong2024robust}. 

Overall, deep learning-based JSCC offers a powerful foundation for semantic communication by jointly optimizing content relevance and channel awareness, moving beyond the bit-wise treatment characteristic of classical approaches.

\subsection{Secure JSCC and Semantic Transmission}

With the rise of semantic communications, security has become a concern, and several works have addressed secure Deep JSCC in the face of eavesdropping.  Early efforts such as DeepJSCEC by Tung and Gündüz~\cite{tung2023deep} integrated DeepJSCC with Learning With Errors LWE-based encryption, achieving resistance to chosen-plaintext attacks while preserving image quality. 
To counter threats from intelligent eavesdroppers, Tang et al.~\cite{tang2025towards, tang2024secure} proposed covert semantic communication using invertible neural networks (INNs) for signal steganography, effectively concealing private information from advanced model inversion attacks.
Beyond covert methods, encryption-compatible architectures have been introduced. Meng et al.~\cite{meng2025secure} introduced a homomorphic encryption (HE)-based framework that enables semantic inference directly on ciphertexts without sacrificing performance.
Additionally, learning-based defenses tackle secrecy-utility trade-offs. Kalkhoran et al.~\cite{kalkhoran2023secure} extended DeepJSCC to scenarios with multiple eavesdroppers using a privacy funnel loss, while Tang et al.\cite{tang2024secure} enhanced robustness through semantic-aware adversarial training. 

These works demonstrate that secure, attack-resilient JSCC can outperform traditional methods under adversarial threats.
However, most existing approaches focus on using JSCC for encryption or eavesdropping prevention. Few studies explicitly address active adversaries such as spoofing or jamming attacks, leaving a gap in the design of robust JSCC systems under intentional disruption.

\subsection{GenAI and Diffusion Model–Enhanced JSCC}

Deep JSCC inherently leverages the structure of autoencoders (AEs) or variational autoencoders (VAEs)~\cite{bourtsoulatze2019deep} to perform joint compression and robust transmission over noisy channels. While early models focused on minimizing distortion, recent works have incorporated generative models to improve perceptual and semantic quality.
Erdemir et al.~\cite{erdemir2023generative} proposed GAN-based schemes such as InverseJSCC and GenerativeJSCC, where a pretrained StyleGAN is used at the receiver to reconstruct high-fidelity images from degraded signals, especially under bandwidth constraints.



Yilmaz et al.~\cite{yilmaz2024high} integrate a denoising diffusion probabilistic model (DDPM) at the receiver to refine the null-space components of DeepJSCC reconstructions. Wu et al.~\cite{wu2024cddm} proposed CDDM, which introduces a channel-aware diffusion process and performs denoising conditioned on channel estimates, significantly improving robustness in varying wireless environments.
Additionally, Wang et al.~\cite{wang2025diffcom} presented DiffCom, which uses the raw received signal as a natural condition for posterior sampling, bypassing the need for explicit CSI and achieving enhanced realism and robustness. Guo et al.~\cite{guo2025diffusion} developed a bandwidth-efficient framework by integrating VAE-based downsampling and upsampling modules into the diffusion pipeline, enabling compatibility with pre-trained diffusion priors under severe channel constraints. Yang et al.~\cite{yang2024diffusion} further proposed DiffJSCC, which fine-tunes latent diffusion with multimodal conditions, including CSI and textual prompts, to guide high-fidelity reconstructions even when the initial JSCC output is severely degraded.

Despite their strong performance, most diffusion-based schemes still suffer from high inference latency due to iterative sampling. To address this, our work introduces a plug-and-play diffusion enhancement module with significantly reduced sampling steps, enabling fast, high-quality reconstruction suitable for communication under adversarial channel perturbations.
\section{System Model}
\vspace{-1mm}

\begin{figure}[htp]
    \centering
    \includegraphics[width= 0.95\linewidth]{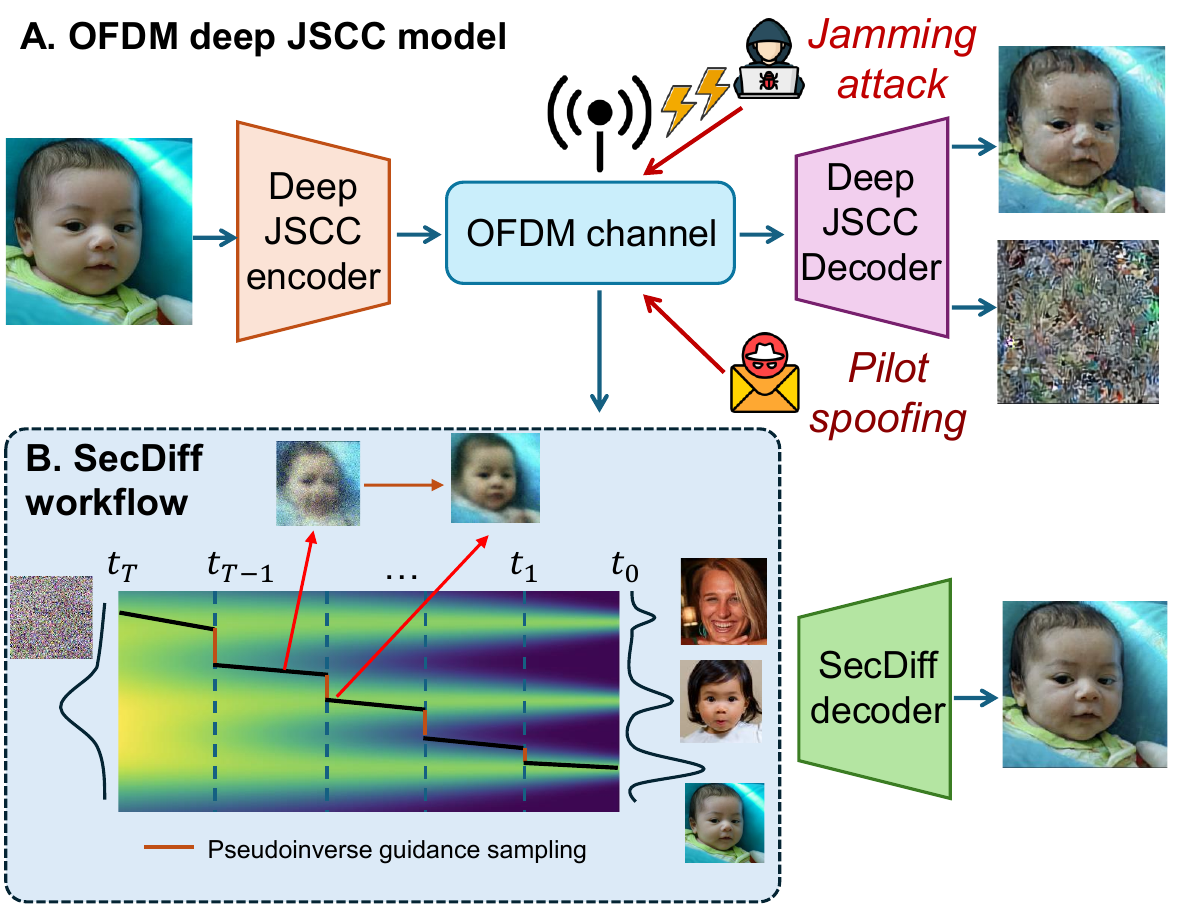}
    \caption{An illustration of SecDiff for secure JSCC over adversarial OFDM channels.
\textit{Part A}: images are transmitted over an OFDM channel using a deep JSCC, but suffer from quality degradation under jamming or pilot spoofing attacks.
\textit{Part B} the proposed SecDiff workflow, where a diffusion-based decoder, guided by pseudoinverse sampling.}
    \label{fig:jscc}
\end{figure}

\subsection{System Overview}

The overall framework of the considered deep JSCC system is illustrated in Fig.~\ref{fig:jscc}. The system comprises an encoder-decoder pair, where the encoder $\mathbf{E}_\theta$ transforms a real-valued source signal $\mathbf{x} \in \mathbb{R}^n$, such as an image or sensor data, into a sequence of complex-valued baseband samples $\mathbf{y} \in \mathbb{C}^m$, which are subsequently transmitted over a wireless channel~\cite{bourtsoulatze2019deep}. 
On the receiver side, the decoder $\mathbf{D}_\phi$ reconstructs the original signal $\mathbf{\hat{x}}$ from the received noisy samples $\mathbf{\hat{y}}$. Here, $\theta$ and $\phi$ denote the trainable parameters of the encoder and decoder neural networks, respectively. Typically, the dimension of the received signal $m$ is smaller than that of the source signal $n$, enabling the encoder to jointly perform source and channel compression. 

The end-to-end training is performed by treating the entire communication process, including channel effects, as a differentiable function. This enables the system to jointly optimize source compression, modulation, and error resilience under a unified deep learning framework. 

\subsection{Channel Model}

Following the framework in~\cite{yang2022ofdm}, we adopt a multipath Rayleigh fading channel model combined with OFDM processing to effectively handle frequency-selective fading in wireless image transmission.



Each multipath component follows an independent Rayleigh distribution $h_l \sim \mathcal{CN}(0, \sigma_l^2)$ for $l = 0, 1, \ldots, L{-}1$, where $L$ denotes the number of resolvable multipath components. The power delay profile follows an exponential decay:
\begin{equation}
\sigma_l^2 = \alpha_l e^{-l / \gamma},
\end{equation}
where $\alpha_l$ is normalized to ensure $\sum_{l=0}^{L-1} \sigma_l^2 = 1$, and $\gamma$ is the delay spread constant.

To mitigate intersymbol interference caused by multipath fading, we employ OFDM with $N$
subcarriers and cyclic prefix length $L_{cp} \geq L-1$ \cite{yang2022ofdm}. Each OFDM frame contains $N_s$ information symbols and $N_p$ pilot symbols for channel estimation.

In the frequency domain, the received pilot $\hat{Y}_p$ and data symbols $\hat{Y}$ are expressed as:
\begin{align}
\hat{Y}_p[i,k] &= H[k] X_p[i,k] + V[i,k], \label{eq:pilot}\\
\hat{Y}[j,k] &= H[k] X[j,k] + W[j,k], \label{eq:data}
\end{align}
where $H[k]$ is the channel frequency response of the
$k$-th subcarrier, $X_p[i,k]$ and $X[j,k]$ represent the transmitted pilot and data symbols, respectively, and
$V[\cdot,\cdot]$ and $W[\cdot,\cdot]$ are noise samples in the frequency domain.

We mitigate the high peak-to-average power ratio (PAPR) of OFDM, which increases power amplifier consumption, by adopting signal clipping~\cite{kim2017novel}. This technique limits the signal amplitude to a threshold determined by the clipping ratio while preserving its phase~\cite{ochiai2002performance}. In our JSCC framework~\cite{yang2022ofdm}, clipping is implemented as a differentiable non-linear activation with power re-normalization, enabling the encoder and decoder to jointly adapt to clipping-induced distortion during end-to-end training.


For channel estimation, we employ per-subcarrier MMSE estimation using the received pilot symbols~\cite{felix2018ofdm}:
\begin{equation}
\hat{H}[k] = \frac{\sum_{i=1}^{N_p} \hat{Y}_p[i,k] X_p^*[i,k]}{N_p + \sigma^2},
\end{equation}
where $X_p[i,k]$ are unit-power pilot symbols known at both transmitter and receiver, and $\sigma^2$ represents the noise power per subcarrier.
Subsequently, we apply MMSE equalization to compensate for channel distortion:
\begin{equation}
Y_{MMSE}[j,k] = \frac{\hat{Y}[j,k] \hat{H}^*[k]}{|\hat{H}[k]|^2 + \sigma^2}.
\end{equation}
This OFDM-based channel model enables the joint source-channel coding system to effectively handle multipath fading while maintaining differentiability for end-to-end optimization~\cite{yang2022ofdm}. 
The channel estimation and equalization operations are incorporated as differentiable layers within the deep learning framework, enabling gradient backpropagation through the entire communication pipeline.

\subsection{Subcarrier Jamming Attack Model}

We consider an active narrowband jamming scenario involving a smart and stealthy adversary capable of selectively degrading deep JSCC-based transmission while avoiding being detected~\cite {esli2006antijamming, lightfoot2010performance, al2016narrow}. 
We consider the adversary possesses precise knowledge of the subcarrier spacing and OFDM symbol timing, enabling it to synchronize its interference with each frame using known preambles. Despite this capability, the jammer is constrained by power limitations or regulatory policies, preventing full-band jamming. 
The jammer targets a subset of OFDM subcarriers to inject interference, with the total number of subcarriers denoted by $N$ and the jammed subset represented as $\mathcal{J} \subseteq {0, 1, \dots, N{-}1}$, where $|\mathcal{J}| = \rho N$ and $0 < \rho < 1$ defines the jamming ratio.
Note that we assume that pilot symbols are left unjammed to maintain essential system functionality~\cite{esli2006antijamming}.

To maximize its disruptive impact under such constraints, the jammer adopts a contiguous jamming strategy, concentrating its power on a continuous block of subcarriers. This focused attack degrades the signal quality on the affected frequencies, inducing partial semantic loss in the transmitted content while keeping the jamming profile sparse and stealthy.

In this scenario, the frequency-domain received signal on subcarrier $k \in \{0, 1, \dots, N{-}1\}$ is modeled as~\cite{esli2006antijamming}:
\begin{equation}
\hat{Y}[j,k] = H[k]X[j,k] + W[j,k] + I[j,k].
\end{equation}
$I[j,k]$ is the jamming interference, defined as:
\begin{equation}
I[j,k] =
\begin{cases}
0, & k \notin \mathcal{J}, \\
J[j,k], & k \in \mathcal{J}.
\end{cases}
\end{equation}
The proposed model is general and can capture arbitrary jamming patterns~\cite{esli2006antijamming}. For analysis in this work, we focus on the case where the jammer generates independent complex Gaussian interference with equal power across all targeted subcarriers.
Specifically, for each jammed subcarrier $k \in \mathcal{J}$, the jamming signal is modeled as:
\begin{equation}
    J[j,k]\sim \mathcal{CN}(0, P_J),
\end{equation}
where $P_J$ denotes the per-subcarrier jamming power.

To quantify the jamming impact, we define the total-band jamming-to-signal ratio (J/R) as~\cite{ma2022jamming}:
\begin{equation}
\text{J/R} 
= 10\log_{10}(\frac{ |\mathcal{J}|\cdot P_J }{ \sum_{k=0}^{N-1} P_{d}[k]}),
\end{equation}
where $|\mathcal{J}| = \rho N$ is the number of jammed subcarriers and $P_{d}[k]$ denotes the transmitted signal power on subcarrier $k$.

\subsection{Pilot Spoofing Attack Model}

We consider a pilot spoofing attack scenario involving a smart adversary aiming to contaminate the channel estimation process in deep JSCC-based OFDM systems~\cite{zhang2014pilot}. Specifically, we assume that the attacker transmits counterfeit pilot symbols that are time- and frequency-aligned with the legitimate pilot transmissions. By leveraging the publicly known structure of standardized pilot sequences, the attacker aims to inject false channel information and compromise the receiver's estimation accuracy.

Under this model, the received pilot symbol on subcarrier $k$ can be expressed as:
\begin{equation}
Y_p[i,k] = H[k]X_p[i,k] + H_a[k]X_s[i,k] + V[i,k],
\end{equation}
where $X_p[i,k]$ is the known pilot symbol, $H[k]$ and $H_a[k]$ denote the channel responses from the legitimate transmitter and the adversary to the receiver, respectively, and $X_s[i,k]$ is the spoofed pilot symbol sent by the adversary.

Unlike subcarrier jamming, which selectively disrupts a subset of data subcarriers, pilot spoofing targets the pilot symbols across all subcarriers, contaminating the entire channel estimation process and causing global degradation. This can result in severe equalization errors and substantial performance loss in JSCC decoding.

\section{The proposed method}

In response to the aforementioned attacks, we propose SecDiff, a plug-and-play diffusion-aided JSCC framework, as illustrated in Fig.~\ref{fig:alg}.
First, to counter subcarrier jamming, SecDiff integrates a power-based detection module with hybrid masked inpainting, enabling robust recovery through pseudoinverse-guided diffusion.
Next, to defend against pilot spoofing, an EM-driven channel refinement mechanism alternates between signal reconstruction and operator estimation, ensuring stable decoding without explicit CSI.
In this section, we detail the proposed framework and its core components.

\subsection{Diffusion Overview}

Diffusion models are a class of generative models that learn to reverse a gradual noise injection process, enabling high-quality sample generation from complex data distributions~\cite{zhao2025supervised}. The key insight is to transform the challenging problem of directly modeling $p_0(x)$ into learning a sequence of simpler denoising operations.

In the variance-preserving (VP) formulation, the diffusion process incrementally perturbs the data while maintaining constant total variance, progressively transforming the original distribution into a tractable Gaussian prior. At each time step, the forward Markov process adds Gaussian noise according to~\cite{ho2020denoising}:
\begin{equation}
q(x_{t+1}|x_t) = \mathcal{N}(x_{t+1}; \sqrt{\alpha_{t+1}}x_t, \beta_{t+1}\mathbf{I}),
\end{equation}
where $\mathbf{I}$ denotes the identity matrix, $\alpha_t = 1 - \beta_t$, and ${\beta_t}_{t=1}^T$ specifies the noise schedule.

By recursively applying this process, we obtain a closed-form expression for the marginal distribution of $x_t$ conditioned on the original clean sample $x_0 \sim p_0(x)$:
\begin{equation}
q(x_t|x_0) = \mathcal{N}(x_t; \sqrt{\bar{\alpha}_t}x_0, (1 - \bar{\alpha}_t)\mathbf{I}),
\end{equation}
where $\bar{\alpha}_t = \prod_{s=1}^t \alpha_s$ denotes the cumulative product of the noise schedule up to step $t$.

Under the variance-preserving condition, the signal term $\sqrt{\bar{\alpha}_t}x_0$ and noise term $(1-\bar{\alpha}_t)$ maintain $\text{Var}[x_t] = \text{Var}[x_0]$ for all $t$. Therefore, as $t \to T$, we have $\bar{\alpha}_T \to 0$, which transforms the data distribution into a standard Gaussian prior:
\begin{equation}
    q(x_T) \approx \mathcal{N}(0, \mathbf{I}).
\end{equation}

The reverse generative process in the VP setting is governed by the following stochastic differential equation (SDE)~\cite{song2020score}:
\begin{equation}
dx = -\frac{1}{2}\beta_t(x + \nabla_x \log p_t(x))dt + \sqrt{\beta_t}d\omega_t,
\end{equation}
where $\nabla_x \log p_t(x)$ is the score function, $\omega_t$ is the standard Wiener process, and $\beta_t$ is the noise schedule. This can be rewritten in terms of the noise prediction formulation~\cite{song2020score}:
\begin{equation}
dx = -\frac{1}{2}\beta_t\left(\frac{x - \sqrt{\bar{\alpha}_t}~\epsilon_\theta(x_t, t)}{\sqrt{1-\bar{\alpha}_t}}\right)dt + \sqrt{\beta_t}d\omega_t,
\end{equation}
where $\epsilon_\theta(x_t, t)$ is a neural network trained to predict the noise component.

The discrete-time sampling procedure of DDPM updates the sample using the learned noise predictor $\epsilon_\theta(x_t, t)$ as follows:
\begin{equation}
\label{eq:ddpm}
x_{t-1} = \frac{1}{\sqrt{\alpha_t}} \left(x_t - \frac{1 - \alpha_t}{\sqrt{1 - \bar{\alpha}_t}}~\epsilon_\theta(x_t, t)\right) + \sigma_t z, \quad z \sim \mathcal{N}(0, \mathbf{I}),
\end{equation}
where $\sigma_t^2 = \beta_t$ is the noise variance at step $t$.

The accelerated variant, known as denoising diffusion implicit models (DDIMs), enables step-skipping for faster sampling by using a non-uniform subsequence of timesteps.
For timesteps set $\{t_1, t_2, \ldots, t_S\}$, where $S$ is the reduced number of sampling steps selected from the original $T$ diffusion steps, the DDIM sampling can be written as~\cite{song2020score}:
\begin{equation}
\label{eq:ddim-noisy}
\begin{split}
x_{t_{i-1}} =& \sqrt{\bar{\alpha}_{t_{i-1}}} \left( \frac{x_{t_i} - \sqrt{1 - \bar{\alpha}_{t_i}} \cdot \epsilon_\theta(x_{t_i}, t_i)}{\sqrt{\bar{\alpha}_{t_i}}} \right) \\
&+ \sqrt{1 - \bar{\alpha}_{t_{i-1}} - \sigma_{t_i}^2} \cdot \epsilon_\theta(x_{t_i}, t_i) + \sigma_{t_i} \cdot z,
\end{split}
\end{equation}
where $\sigma_{t_i}^2$ controls the amount of stochasticity, and can be set to $0$ for deterministic sampling (DDIM), or to $\beta_{t_i}$ for fully stochastic sampling, and $z \sim \mathcal{N}(0, \mathbf{I})$ is standard Gaussian noise. This formulation allows jumping directly between non-consecutive timesteps, significantly reducing the number of denoising steps required from $T$ to $S$ while maintaining sample quality.

The central challenge of diffusion models is learning the score function $s_\theta = \nabla_x \log p_t(x)$ or, equivalently, the noise prediction function $\epsilon_\theta(x_t, t)$. This is addressed through the denoising objective~\cite{ho2020denoising}:
\begin{equation}
    \mathcal{L}(\theta) = \mathbb{E}_{t,x_0,\epsilon} \left[\|\epsilon - \epsilon_\theta(\sqrt{\bar{\alpha}_t}x_0 + \sqrt{1-\bar{\alpha}_t}\epsilon, t)\|^2\right],
\end{equation}
where $\epsilon \sim \mathcal{N}(0, \mathbf{I})$ is the noise sample and the expectation is taken over timesteps $t \sim \text{Uniform}(1, T)$.
Sample generation proceeds by solving the reverse SDE starting from pure noise $x_T \sim \mathcal{N}(0, \mathbf{I})$ and integrating backward in time using numerical solvers such as the DDPM sampler or DDIM sampler in Eqs. \eqref{eq:ddpm} and \eqref{eq:ddim-noisy}, respectively.

\subsection{Inverse-guided Sampling}

In inverse problems, we seek to recover an unknown signal $x_0 \in \mathbb{R}^n$ from indirect measurements $y \in \mathbb{R}^m$. The measurement process is typically modeled as:
\begin{equation}
\label{eq:inverse}
    y = \mathcal{H}(x_0) + z,
\end{equation}
where $\mathcal{H}: \mathbb{R}^n \to \mathbb{R}^m$ is a known measurement operator and $z \sim \mathcal{N}(0, \sigma_y^2)$ represents measurement noise. Classical approaches often suffer from ill-posedness, requiring strong regularization assumptions that may not hold in practice.

Diffusion models offer a principled approach to inverse problems by leveraging learned priors. A central idea is to sample from the posterior distribution $p_t(x_t|y)$ during the reverse diffusion process. By applying Bayes' rule, the conditional score function can be decomposed as~\cite{chung2022diffusion}:
\begin{equation}
    \nabla_{x_t} \log p_t(x_t|y) = \nabla_{x_t} \log p_t(x_t) + \nabla_{x_t} \log p_t(y|x_t),
\end{equation}
where the first term corresponds to the score of the diffusion prior and can be directly estimated by a pre-trained score network.
The second term, often referred to as the data consistency or guidance term, involves computing the likelihood $p_t(y|x_t)$, which requires marginalizing over the clean data:
\begin{equation}
    p_t(y|x_t) = \int p(x_0|x_t)p(y|x_0)~dx_0.
\end{equation}

Effectively estimating this intractable marginalization is the key challenge in applying diffusion models to inverse problems, as it directly influences the fidelity of posterior sampling and the overall quality of the reconstructed solution.

\subsection{Pseudoinverse Guidance}

Pseudoinverse guidance addresses this challenge via a principled approximation strategy~\cite{song2023pseudoinverse}. The core idea is to approximate the intractable posterior $p_t(x_0|x_t)$ with a Gaussian distribution:
\begin{equation}
    p_t(x_0|x_t) \approx \mathcal{N}(\hat{x}_t, r_t^2 \mathbf{I}),
\end{equation}
where $r_t$ denotes a time-dependent variance parameter that captures the uncertainty in the one-step denoising process. The posterior mean $\hat{x}_t$ is estimated using Tweedie’s formula~\cite{kim2022noise}, serving as the MMSE estimator:
\begin{equation}
    \hat{x}_t = \mathbb{E}[x_0|x_t] = x_t + \sigma_t^2 \nabla_{x_t} \log p_t(x_t) \approx x_t + \sigma_t^2 s_\theta(x_t, \sigma_t),
\end{equation}
where $s_\theta$ is the learned score function.

For linear measurements $H\in\mathbb{R}^{m\times n}$ with Gaussian noise, this approximation leads to
\begin{equation}
    p_t(y|x_t) \approx \mathcal{N}(H\hat{x}_t, r_t^2 HH^T + \sigma_y^2 \mathbf{I}).
\end{equation}
Thus, the resulting pseudoinverse guidance term becomes~\cite{song2023pseudoinverse}:
\begin{equation}
\begin{split}
\nabla_{x_t} \log p_t(y|x_t) \approx \left[(y - H\hat{x}_t)^T \left(r_t^2 HH^T + \sigma_y^2 \mathbf{I}\right)^{-1} H \frac{\partial \hat{x}_t}{\partial x_t}\right]^T.
\end{split}
\end{equation}
This is computed as a vector-Jacobian product through automatic differentiation, requiring backpropagation only through the score network~\cite{song2023pseudoinverse}.

For noiseless measurements, the expression simplifies to
\begin{equation}
    \nabla_{x_t} \log p_t(y|x_t) \approx r_t^{-2} \left[(H^\dagger y - H^\dagger H\hat{x}_t)^T \frac{\partial \hat{x}_t}{\partial x_t}\right]^T,
\end{equation}
where $H^\dagger = H^T(HH^T)^{-1}$ is the Moore-Penrose pseudoinverse of $H$.

A key advantage of pseudoinverse guidance is its extension to non-linear and non-differentiable measurement operators~\cite{song2023pseudoinverse}. For a general measurement function $\mathcal{H}$, we define a generalized pseudoinverse $\mathcal{H}^\dagger: \mathbb{R}^m \to \mathbb{R}^n$ such that 
\begin{equation}
\label{eq:pseudoinverse}
    \mathcal{H}(\mathcal{H}^\dagger(\mathcal{H}(x))) = \mathcal{H}(x),
\end{equation}
for all $x$. The guidance term then becomes
\begin{equation}
\label{eq:guidance1}
    \nabla_{x_t} \log p_t(y|x_t) \approx r_t^{-2} \left[(\mathcal{H}^\dagger(y) - \mathcal{H}^\dagger(\mathcal{H}(\hat{x}_t)))^T \frac{\partial \hat{x}_t}{\partial x_t}\right]^T.
\end{equation}


This framework can be naturally extended to deep JSCC systems operating over OFDM-based multipath fading channels. To reflect the realistic transmission process in OFDM-based systems, we define an effective time-domain channel operator $\mathcal{H}_{\mathrm{ofdm}}(\cdot)$ that encapsulates all physical-layer transformations, including OFDM modulation, multipath fading, and additive noise. The received signal is thus modeled as:
\begin{equation}
    \hat{\mathbf{y}} = \mathcal{H}_{\mathrm{ofdm}}(\mathbf{E}_\theta(\mathbf{x})),
\end{equation}
where $\mathbf{E}_\theta(\cdot)$ is the deep JSCC encoder, and $\hat{\mathbf{y}}$ is the corrupted time-domain observation at the receiver.
The JSCC decoder $\mathbf{D}_\phi$ then aims to recover the source signal by learning:
\begin{equation}
    \mathbf{D}_\phi(\hat{\mathbf{y}}) \approx \mathbf{x},
\end{equation}
minimizing the reconstruction loss over the end-to-end transmission path.



This objective aligns precisely with the definition of the generalized pseudoinverse in Eq. \eqref{eq:pseudoinverse}, where the decoder acts as the pseudoinverse of the deep JSCC encoder:
\begin{equation}
    \mathbf{D}_\phi \approx ( \mathcal{H}_{\mathrm{ofdm}}(\mathbf{E}_\theta))^{\dagger}.
\end{equation}
Based on this, the guidance term for deep JSCC can be written as:
\begin{equation}
\label{eq:jsccguidance}
    g = r_t^{-2} \left[(\mathbf{D}_\phi(\mathbf{y}) - \mathbf{D}_\phi(\mathcal{H}_{\mathrm{ofdm}}(\mathbf{E}_\theta(\mathbf{\hat{x}}_t)))^T \frac{\partial \mathbf{\hat{x}}_t}{\partial \mathbf{x}_t}\right]^T.
\end{equation}

The guided DDIM sampling update from time $t_i$ to $t_{i-1}$ with effective channel operator becomes:
\begin{equation}
\label{eq:ddim-guided}
\begin{split}
\mathbf{x}_{t_{i-1}} =& \sqrt{\bar{\alpha}_{t_{i-1}}} \left( \frac{\mathbf{x}_{t_i} - \sqrt{1 - \bar{\alpha}_{t_i}} \cdot \epsilon_\theta(\mathbf{x}_{t_i}, t_i)}{\sqrt{\bar{\alpha}_{t_i}}} \right) \\
&+ \sqrt{1 - \bar{\alpha}_{t_{i-1}} - \sigma_{t_i}^2} \cdot \epsilon_\theta(\mathbf{x}_{t_i}, t_i) + \sigma_{t_i} \cdot \mathbf{z} + r_{t_i}^2 \cdot g.
\end{split}
\end{equation}
The adaptive scaling $r_{t_i}^2$ increases guidance strength at early steps and reduces it near convergence, enhancing stable and accurate reconstructions. 

This formulation effectively combines the generative capability of diffusion models with principled measurement consistency, enabling plug-and-play inference in degraded or adversarial environments.

\begin{figure*}[htp]
    \centering
    \includegraphics[width= 0.95\linewidth]{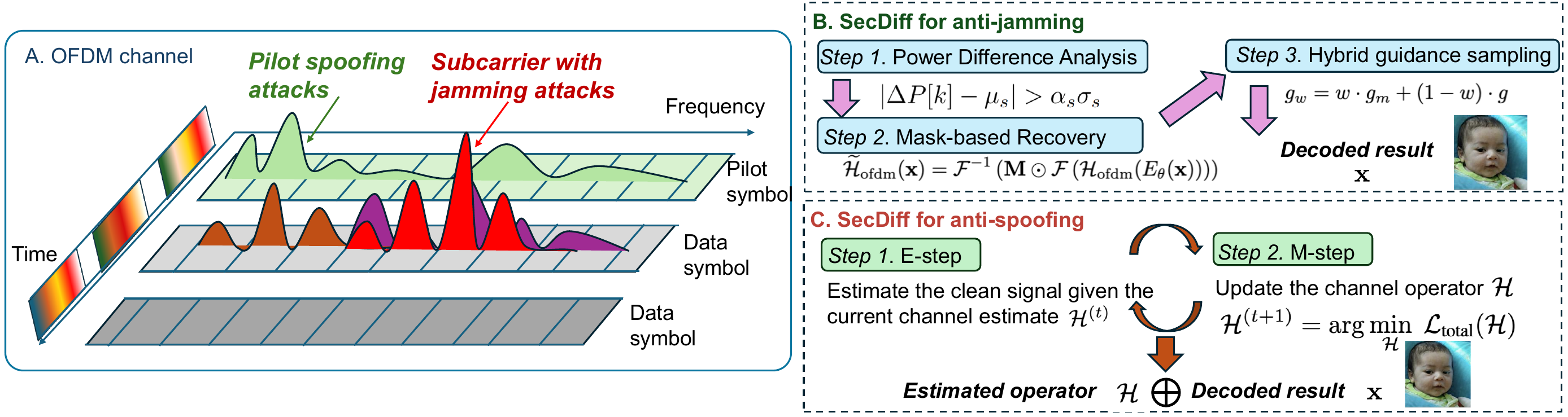}
    \caption{An overview of SecDiff for secure JSCC over adversarial OFDM channels.
\textit{Part A}: adversarial OFDM channels suffer from pilot spoofing and subcarrier jamming.
\textit{Part B} SecDiff detects jammed subcarriers, restores them via inpainting, and performs hybrid diffusion decoding.
\textit{Part C} SecDiff adopts an EM-style refinement of the channel operator for robust signal reconstruction.}
    \label{fig:alg}
\end{figure*}

\subsection{Subcarrier Jamming Attack}


To mitigate jamming attacks, we propose a diffusion-enhanced reconstruction method guided by a subcarrier-aware masking mechanism, as shown in Fig. \ref{fig:alg} \textit{Part B}.



\subsubsection{Jamming Detection via Power Difference Analysis}

To combat subcarrier jamming, we first need to identify which subcarriers are compromised. We utilize a power difference-based detection method that exploits the statistical properties of pilot symbols and data symbols~\cite{han2008ofdm}.




First, we compute the average power spectrum on each subcarrier by averaging over the frame, packet, and symbol dimensions. Let $P_p[k]$ and $P_d[k]$ denote the resulting average power of pilot and data signals on subcarrier $k$, respectively.
Next, we compute the power difference between the data and pilot signals on each subcarrier as follows:
\begin{equation}
\Delta P[k] = P_d[k] - P_p[k],
\end{equation}
where $\Delta P[k]$ captures the excess energy potentially caused by subcarrier jamming.



To identify anomalous subcarriers, we estimate the global mean $\mu$ and standard deviation $\sigma$ of $\Delta P[k]$ over all $N$ subcarriers:
\begin{equation}
\mu_s = \frac{1}{N} \sum_{k=1}^{N} \Delta P[k], \quad \sigma_s = \sqrt{\frac{1}{N-1} \sum_{k=1}^{N} (\Delta P[k] - \mu_s)^2}.
\end{equation}
A subcarrier $k$ is flagged as jammed if its power difference exceeds a threshold:
\begin{equation}
\label{eq:var}
|\Delta P[k]-\mu_s| >  \alpha_s \sigma_s,
\end{equation}
where $\alpha_s$ is a tunable sensitivity parameter. In this work, we set $\alpha_s = 2$ following the $2\sigma$ rule, which ensures that approximately 95\% of normal data falls within the threshold~\cite{huber2018logical}.

The proposed method detects subcarriers exhibiting statistically significant power deviations. Although it does not explicitly differentiate between jamming and background noise, it effectively captures anomalous power patterns that may correspond to potential attacks or environmental anomalies. Mild jamming with small power variations may occasionally evade detection, leading to missed detections. Generally, such missed detections usually have a limited impact on the overall transmission quality.








\subsubsection{Inpainting-based Recovery}

Once jammed or anomalous subcarriers are identified, we model the recovery problem as an inpainting task~\cite{chung2022diffusion}. We construct a binary mask $\mathbf{M} \in \{0,1\}^M$ where:
$$\mathbf{M}[k] = \begin{cases}
0, & \text{if subcarrier } k \text{ is jammed} \\
1, & \text{if subcarrier } k \text{ is clean}.
\end{cases}$$
Although $\mathbf{M}$ operates in the frequency domain, its impact is equivalent to a structured degradation in the time domain. We define a masked time-domain transmission operator:
\begin{equation}
\widetilde{\mathcal{H}}_{\mathrm{ofdm}}(\mathbf{x}) = \mathcal{F}^{-1} \left( \mathbf{M} \odot \mathcal{F} \left( \mathcal{H}_{\mathrm{ofdm}}(E_\theta(\mathbf{x})) \right) \right),
\end{equation}
where $\mathcal{F}$ and $\mathcal{F}^{-1}$ denote the forward and inverse discrete Fourier transform (DFT), respectively, $\mathcal{H}_{\mathrm{time}}$ represents the original unmasked OFDM channel operator, and $\odot$ denotes element-wise multiplication. 
The corresponding approximate pseudoinverse is:
\begin{equation}
\widetilde{\mathcal{H}}^{\dagger}_{\mathrm{ofdm}}(\hat{\mathbf{y}}) = \mathbf{D}_\phi\left( \mathcal{F}^{-1} \left( \mathbf{M} \odot \mathcal{F}(\hat{\mathbf{y}}) \right) \right),
\end{equation}
where $\hat{\mathbf{y}}$ is the received time-domain signal, and the masked inverse-DFT simulates the removal of jammed subcarriers.

This formulation transforms the frequency-domain inpainting problem into a time-domain inverse reconstruction task compatible with the JSCC decoder input, allowing us to perform diffusion-guided recovery directly on time-domain observations.




Using the pseudoinverse guidance framework, the guidance term becomes:
\begin{equation}
\begin{split}
        g_m = r_t^{-2} \left[(\widetilde{\mathcal{H}}^{\dagger}_{\mathrm{ofdm}}(\hat{\mathbf{y}})) - \widetilde{\mathcal{H}}^{\dagger}_{\mathrm{ofdm}}(\widetilde{\mathcal{H}}_{\mathrm{ofdm}}(\mathbf{x}))
        )^T \frac{\partial \hat{\mathbf{x}}_t}{\partial \mathbf{x}_t}\right]^T.
\end{split}
\end{equation}

However, since the encoder and decoder did not take the mask into consideration during training, the pseudoinverse constraint $\widetilde{\mathcal{H}}_{\mathrm{ofdm}}(\widetilde{\mathcal{H}}^{\dagger}_{\mathrm{ofdm}}(\widetilde{\mathcal{H}}_{\mathrm{ofdm}}(\cdot))) = \widetilde{\mathcal{H}}_{\mathrm{ofdm}}(\cdot)$ may lose efficacy. To address this, we introduce a weighted hybrid guidance mechanism that interpolates between masked and unmasked terms:
\begin{equation}
\label{eq:hybrid_guidance}
g_w = w\cdot g_m+ (1-w)\cdot g
\end{equation}
where $w \in [0,1]$ is a tunable weight that controls the influence of the masked pseudoinverse guidance versus the unmasked reconstruction guidance.

This weighted formulation ensures robustness across varying jamming intensities: when most subcarriers are jammed, lower $w$ allows the model to leverage global structure; when fewer are jammed, higher $w$ emphasizes consistency with clean observations. The impact of this hyperparameter is further analyzed in our experiments.

\subsection{Pilot Spoofing Attack}

\subsubsection{EM Algorithm with Reconstruction Error Guidance}

Under pilot spoofing attacks, the CSI becomes unreliable due to adversarial contamination of the pilot symbols. As a result, the effective forward operator $\mathcal{H}_{\mathrm{ofdm}}$ can no longer be assumed to be known. This renders the reconstruction problem ill-posed, as both the transmitted signal and the transformation operator are unknown.

To address this challenge, we propose an EM-based pseudoinverse-guided diffusion framework that jointly estimates the latent signal and the unknown channel operator, as shown in Fig. \ref{fig:alg} \textit{Part C}. Specifically, we treat the forward process as a blind inverse problem:
\begin{equation}
    \hat{\mathbf{y}} = \mathcal{H}(\mathbf{x}),
\end{equation}
where $\mathcal{H}$ is an unknown corrupted operator.



We employ an EM algorithm that alternates between~\cite{moon1996expectation}:
\begin{itemize}
    \item \textbf{E-step:} Estimate the clean signal given the current channel estimate $\mathcal{H}^{(t)}$ via Eq. \eqref{eq:ddim-guided}.
    \item \textbf{M-step:} Update the channel operator $\mathcal{H}$ based on the denoised signal samples.
\end{itemize}

To estimate the equivalent channel operator $\mathcal{H}$, the M-step refines $\mathcal{H}$ by minimizing a regularized loss function:
\begin{equation}
\mathcal{H}^{(t+1)} = \arg\min_{\mathcal{H}} \; \mathcal{L}_{\text{total}}(\mathcal{H}),
\end{equation}
where
\begin{equation}
\mathcal{L}_{\text{total}}(\mathcal{H}) = \mathcal{L}_{\text{recon}}(\mathcal{H}) + \lambda_2 \, \mathcal{R}_{\text{prior}}(\mathcal{H}).
\end{equation}

The first term $\mathcal{L}_{\text{recon}}$ enforces measurement consistency between the observed signal and the forward model:
\begin{equation}
\mathcal{L}_{\text{recon}}(\mathcal{H}) = \mathbb{E}_{\mathbf{x} \sim q(\mathbf{x}|\mathbf{y}, \mathcal{H}^{(t)})} \left[ \left\| \mathbf{y} - \mathcal{H}(\mathbf{E}_\theta(\mathbf{x})) \right\|_2^2 \right].
\end{equation}


Since the time-domain channel impulse response $\mathbf{h}_t$ follows a circularly symmetric complex Gaussian distribution, we adopt a Gaussian prior for regularization:
 \begin{equation}
\mathcal{R}_{\text{prior}}(\mathcal{H}) = - \log p(\mathbf{h}) = \sum_{l=0}^{L-1} \frac{|h_l|^2}{\sigma_l^2} + \text{const},
\end{equation}
where $\mathbf{h}=[h_0,h_1,\ldots,h_{L-1}]$ denotes the time-domain multipath channel impulse response.
Since the exact power delay profile is unknown, we assume an equal-power prior by setting $\sigma_l^2 = 1/L$ for all paths.

In practice, we adopt gradient-based optimization to update $\mathcal{H}$:
\begin{equation}
\mathcal{H}^{(t+1)} \leftarrow \mathcal{H}^{(t)} - \eta \nabla_{\mathcal{H}} \mathcal{L}_{\text{total}}(\mathcal{H}),
\end{equation}
where $\eta$ is the learning rate.
This formulation enables structured channel refinement that balances data fidelity and prior conformity.

This joint approach effectively combines the generative power of diffusion models with the parameter estimation capabilities of EM algorithms, enabling robust recovery of clean pilot symbols even in the presence of sophisticated spoofing attacks.

\section{Experimental Results}

\subsection{Implementation Details}

\subsubsection{Dataset and Evaluation Metrics}

To evaluate the performance and robustness of our proposed secure JSCC framework under adversarial conditions, we conduct experiments on the FFHQ 256×256 dataset~\cite{karras2019style}, which consists of high-quality facial images suitable for perceptual evaluation in semantic transmission tasks. We select 300 validation images as the test set for each experiment.

To comprehensively evaluate the proposed method, we employ several categories of image quality assessment (IQA) metrics, including 
PSNR, MS-SSIM~\cite{hore2010image}, LPIPS~\cite{zhang2018perceptual}, DISTS~\cite{ding2020iqa}, and FID~\cite{wang2025generative}.
For all metrics, we report mean values across the test set. Arrows $\uparrow$ and $\downarrow$ indicate whether higher or lower values indicate better performance, respectively.

\subsubsection{Benchmark Methods}

As our goal is to develop a plug-and-play secure decoding module for adversarial wireless environments, we focus on end-to-end JSCC frameworks and exclude traditional source-channel separation schemes from our comparison.

To demonstrate broad compatibility and fairness, we integrate our secure decoder into the widely used DeepJSCC framework~\cite{yang2022ofdm}, which serves as the base encoder. For comparison, we also include Diffcom~\cite{wang2025diffcom}, a recent state-of-the-art diffusion-based JSCC decoder that supports plug-and-play integration and does not require explicit channel estimation.

To ensure a comprehensive and fair evaluation, we consider several variants of the baseline method DiffCom~\cite{wang2025diffcom}, each tailored to different settings:
\begin{itemize}
\item DiffCom-n: the original version with full denoising steps;

\item DiffCom-f: an accelerated variant using adaptive step scheduling;

\item DiffCom-s: a configuration aligned with our method, sharing the same number of denoising steps for direct comparison;

\item DiffCom-b: a blind channel estimation variant for robustness evaluation under unknown channels.
    
\end{itemize}

All models are evaluated using their official pretrained weights and open-source implementations\footnote{https://github.com/wsxtyrdd/diffcom}. In jamming and spoofing scenarios, we fix the channel parameters across methods to isolate the impact of the decoder and its recovery strategy.

\subsubsection{Channel and Diffusion Configuration}

We simulate a multipath Rayleigh fading channel consisting of $L = 8$ paths with an exponential power delay profile characterized by a decay constant of $r = 4$. The OFDM system is configured with an FFT size of $N = 256$ subcarriers and a cyclic prefix of length $L_{\text{cp}} = 10$. Pilot symbols are allocated such that one pilot is placed per subcarrier along the time axis.

For diffusion-based components, we use pretrained DDPM models for the FFHQ dataset with 1000-step noise schedules~\cite{choi2021ilvr}. To reduce sampling time, we apply DDIM sampling with step-skipping and employ our proposed SecDiff.

All experiments are implemented in Python using the PyTorch framework. The server is equipped with Ubuntu 22.04, powered by an Intel(R) Xeon(R) Silver 4410Y 12-core processor and an NVIDIA RTX A6000 GPU.




\begin{figure*}[t]
\centering
\begin{subfigure}{.32\textwidth}
  \centering
  \includegraphics[width=0.90\linewidth]{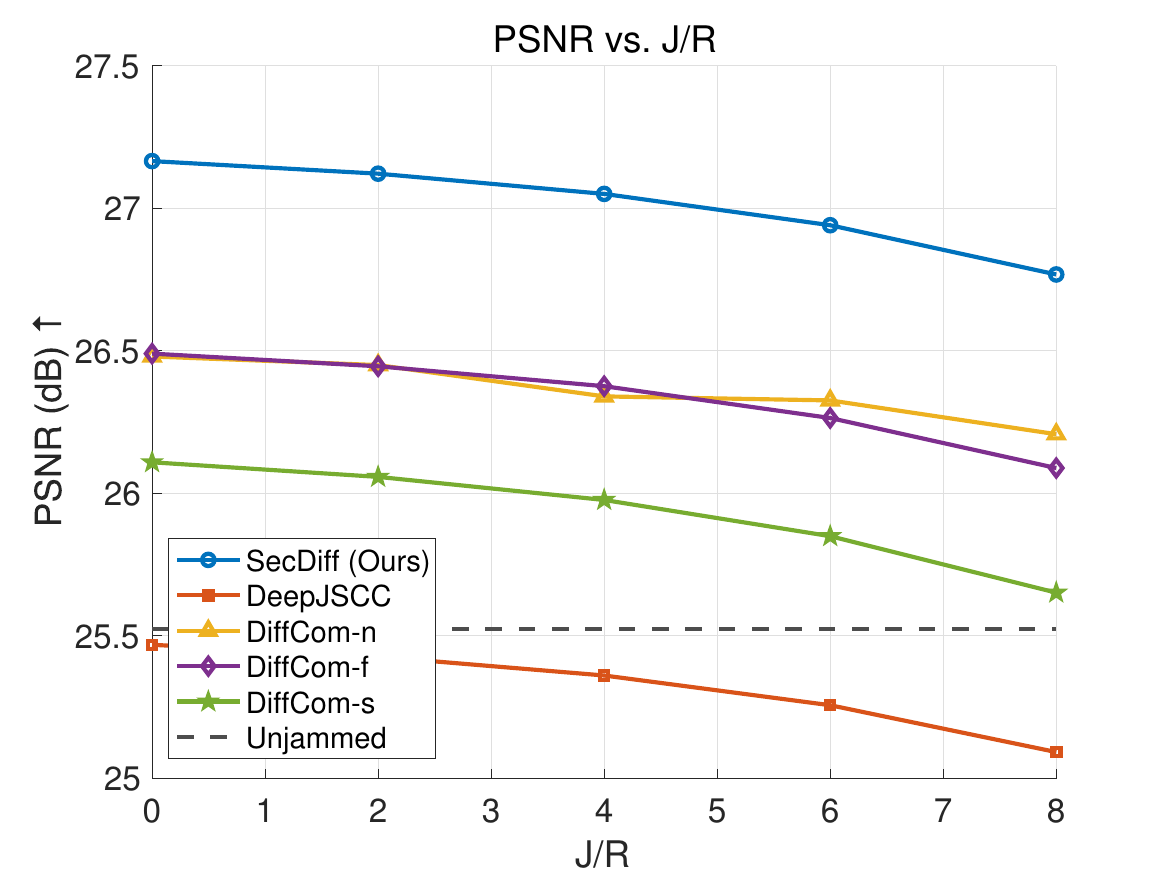}
  \caption{PSNR}
  \label{fig:sub1}
\end{subfigure}
\hfill
\begin{subfigure}{.32\textwidth}
  \centering
  \includegraphics[width=0.90\linewidth]{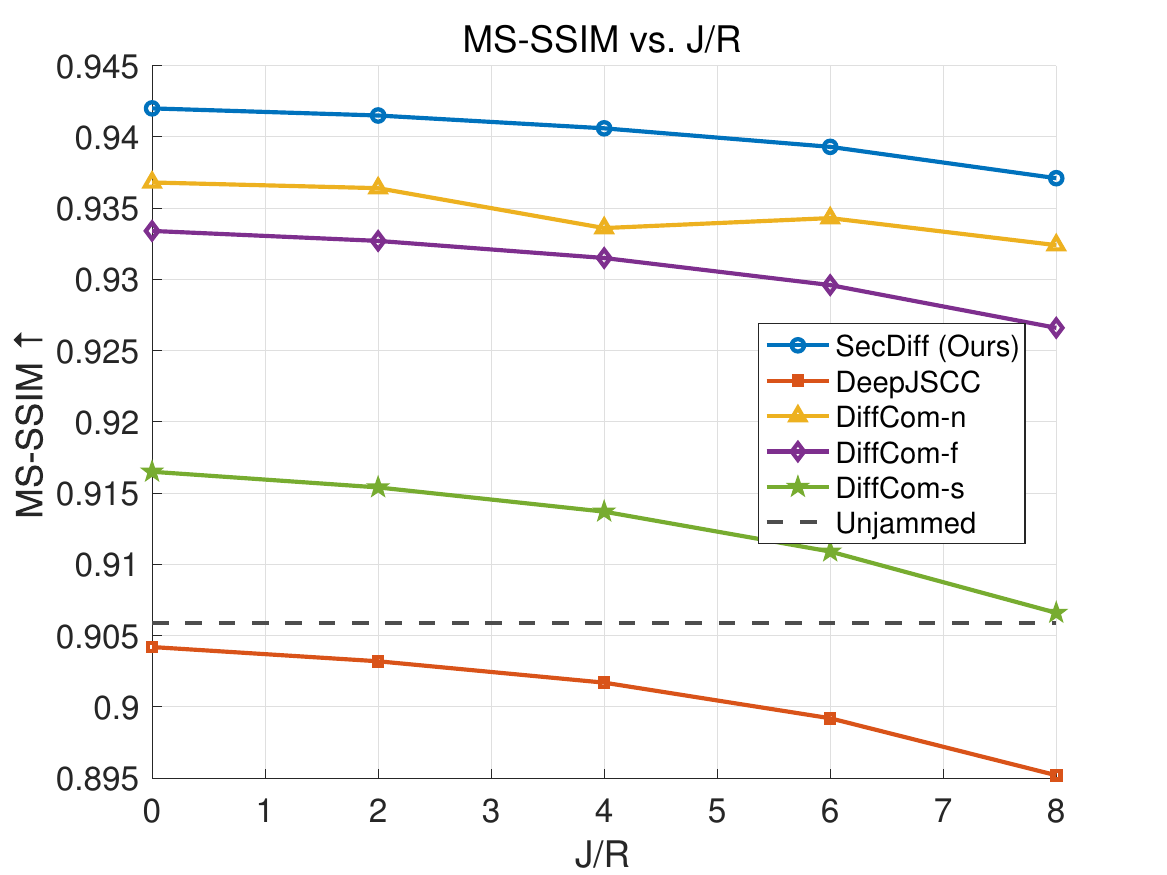} 
  \caption{MS-SSIM}
  \label{fig:sub2}
\end{subfigure}%
\hfill
\begin{subfigure}{.32\textwidth}
  \centering
  \includegraphics[width=0.90\linewidth]{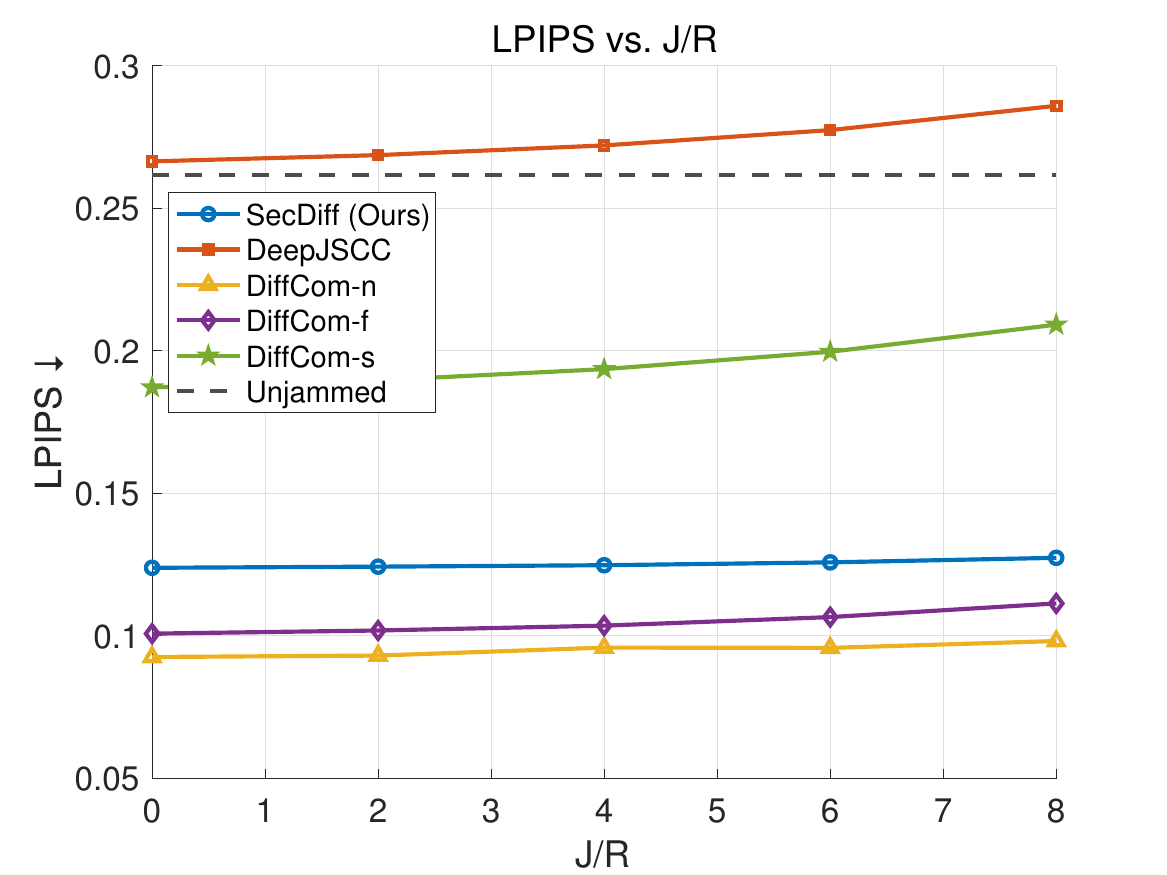}
  \caption{LPIPS}
  \label{fig:sub3}
\end{subfigure}
\vspace{0.4em}
\begin{subfigure}{.32\textwidth}
  \centering
  \includegraphics[width=0.90\linewidth]{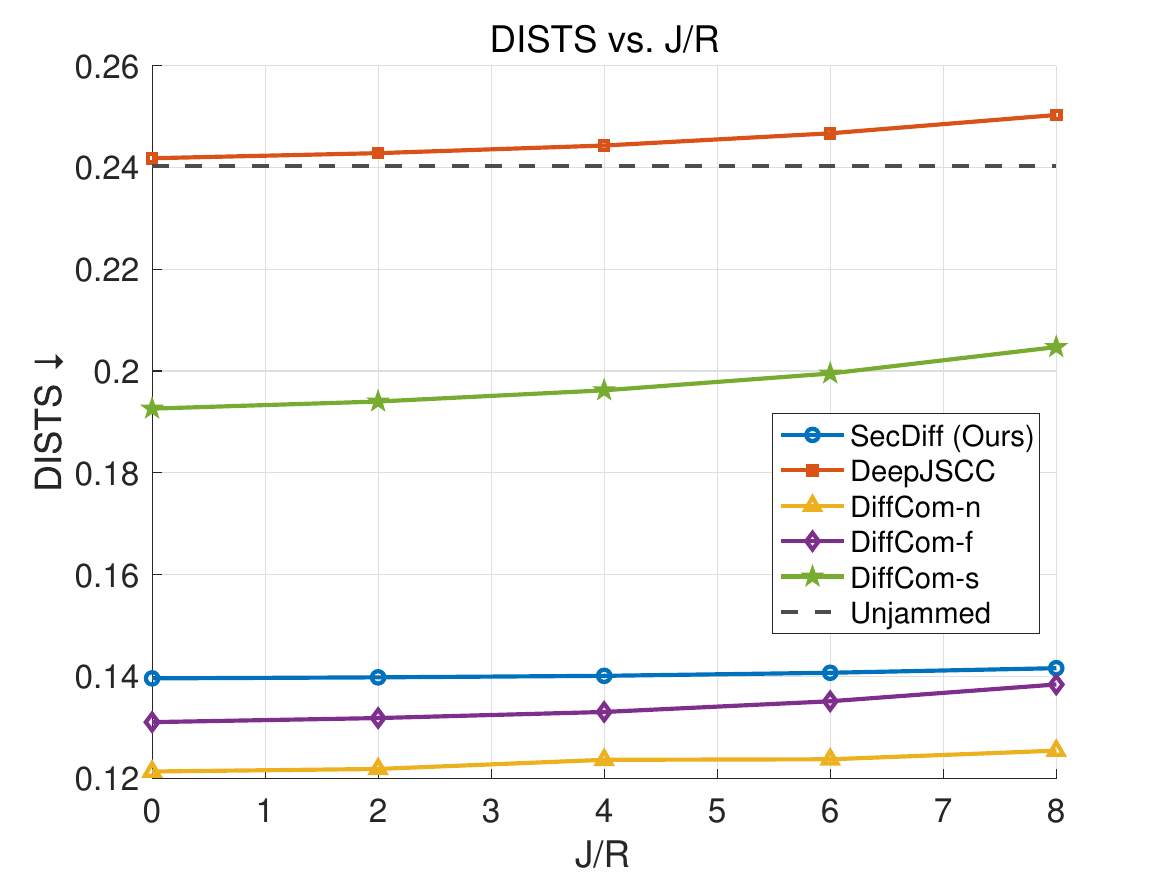}
  \caption{DISTS}
  \label{fig:sub4}
\end{subfigure}
\hspace{0.05\textwidth} 
\begin{subfigure}{.32\textwidth}
  \centering
  \includegraphics[width=0.90\linewidth]{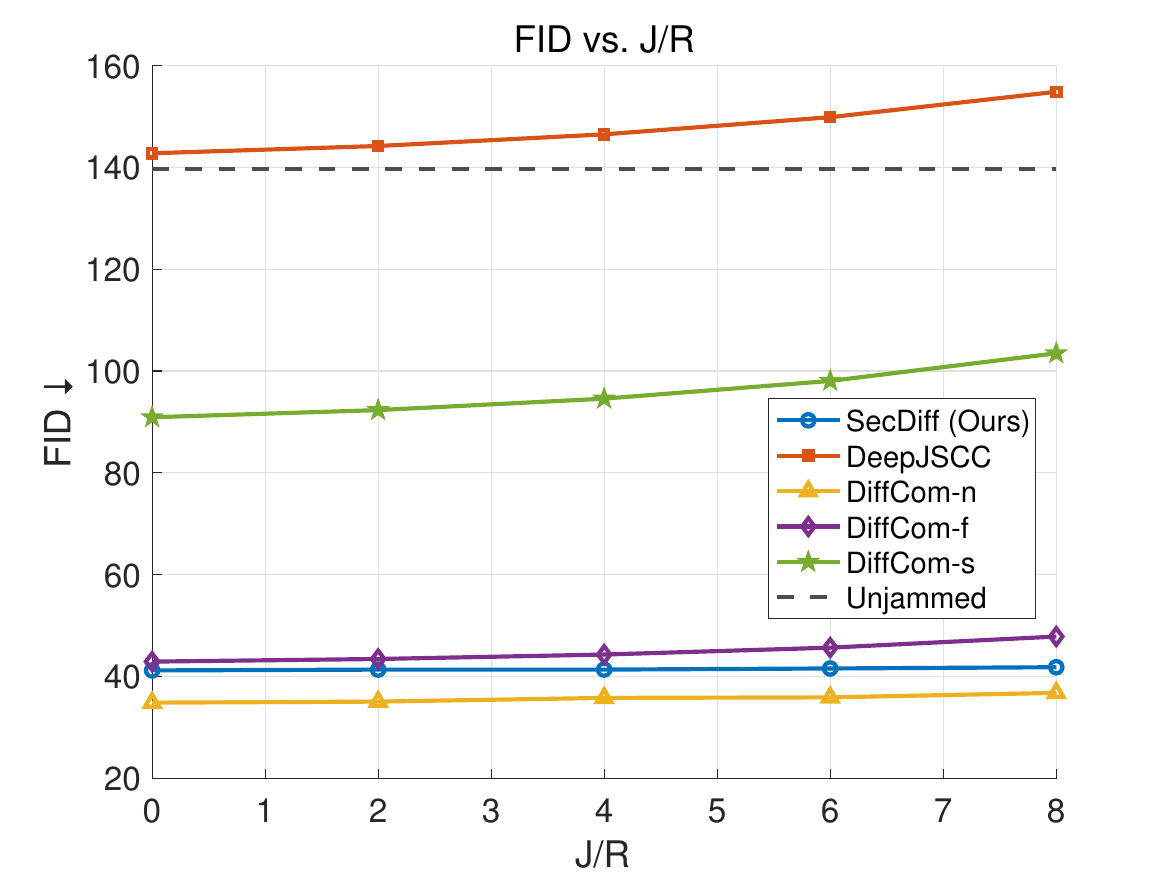}
  \caption{FID}
  \label{fig:sub5}
\end{subfigure}
\caption{Quantitative evaluation of image quality under different J/R levels with fixed attack ratio $\rho=0.2$ and $\text{SNR} = 10~\mathrm{dB}$, using five IQA metrics.}
    \label{fig:five_metrics}
\end{figure*}

\begin{figure*}[htp]
    \centering
    \includegraphics[width= 0.90\linewidth]{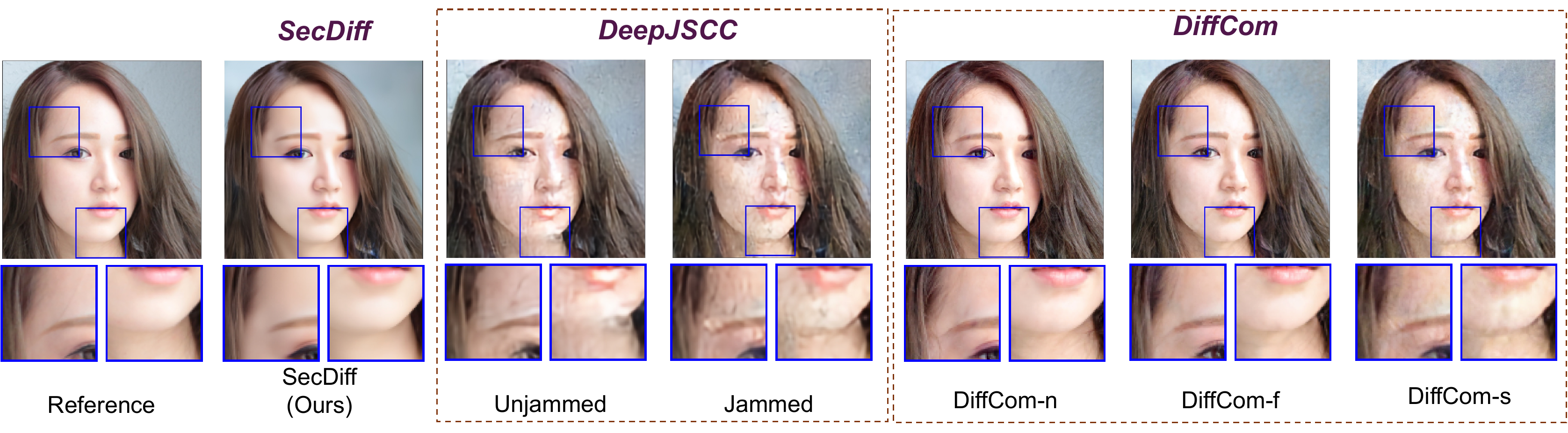}
    \caption{Qualitative comparison of reconstructed images under adversarial jamming with $\rho=0.2$ and $\text{J/R}=8~\mathrm{dB}$. 
    }
    \label{fig:ex1}
\end{figure*}

\subsection{Jamming Defense}

We conduct a comprehensive evaluation of different JSCC schemes under adversarial OFDM settings, where the signal-to-noise ratio (SNR) is fixed at $10~\mathrm{dB}$ and 20\% of subcarriers are jammed, i.e., $\rho = 0.2$. The J/R is gradually increased from $0$ to $8~\mathrm{dB}$ to simulate increasingly hostile conditions, with a constant coding bit rate (CBR) of $1/48$.

\subsubsection{Quantitative Comparison}

Fig.~\ref{fig:five_metrics} illustrates the quantitative evaluation of image quality under various J/R levels ranging from $0$ to $8~\mathrm{dB}$, measured by five IQA metrics. The baseline DeepJSCC exhibits a consistent degradation across all metrics as J/R increases, confirming its vulnerability to frequency-domain attacks and its inability to suppress structured interference or recover corrupted semantic content.
In contrast, the proposed SecDiff, which integrates a diffusion-based decoder with 50 denoising steps, consistently outperforms DeepJSCC across all J/R levels. At $\text{J/R} = 8~\mathrm{dB}$, SecDiff achieves a PSNR of $26.77~\mathrm{dB}$, representing a 6.7\% improvement over DeepJSCC. For LPIPS and DISTS, SecDiff yields scores of 0.1273 and 0.1416, achieving 55.5\% and 43.4\% gains, respectively. Additionally, SecDiff attains a 4.7\% higher MS-SSIM and a 73\% lower FID, demonstrating superior preservation of both low-level fidelity and high-level semantic features.

We further compare SecDiff with three variants of DiffCom. 
The full-step DiffCom-n achieves slightly better perceptual metrics but requires an inference time of approximately $55$~s per image, rendering it impractical for real-time deployment. DiffCom-f reduces runtime to around $14$~s by employing SNR-based adaptive step scheduling~\cite{wang2025diffcom}. However, this speedup comes at the cost of perceptual quality, with 14.4\% higher FID and 12.3\% higher LPIPS compared to SecDiff. DiffCom-s, configured to use the same 50 denoising steps as SecDiff, offers comparable runtime around 3s, but still underperforms than SecDiff. Specifically, at $\text{J/R} = 8~\mathrm{dB}$, its PSNR is 4.4\% lower, LPIPS 39.1\% higher, DISTS 30.8\% higher, MS-SSIM 3.4\% lower, and FID 59.6\% worse than SecDiff. These results demonstrate that merely reducing the step count, as in DiffCom-s, is insufficient to achieve robustness. The noise-aware sampling and variance-adjusted guidance in SecDiff provide effective protection against adversarial perturbations while preserving computational efficiency.

In summary, SecDiff secures substantial improvements in all IQA metrics but also maintains a low inference latency, achieving a favorable trade-off between robustness, quality, and runtime. This balance underscores its practicality for real-time secure semantic communication under adversarial conditions.

\begin{table*}[t]\scriptsize
\centering
\caption{Performance comparison under jamming attacks with varying parameters at $\text{J/R} = 8~\mathrm{dB}$.}
\begin{tabular}{lccccccccc}
\toprule
\textbf{Method} & $\rho$ & $n_{Step}$ & $w$  & PSNR$\uparrow$ & MS-SSIM$\uparrow$ & LPIPS$\downarrow$ & DISTS$\downarrow$ & FID$\downarrow$ & Time (s) \\
\midrule
DeepJSCC & 0.2 & / & / & 25.0914 & 0.8952
 & 0.2859 & 0.2503  & 154.8608 & $0.2949 \bm{\pm} 0.0095$ \\
\midrule
DiffCom-n & 0.2 & 1000 & / & 26.2067 & 0.9324 & \textbf{0.0981} & \textbf{0.1254} & \textbf{36.7508} & $55.4886 \bm{\pm} 0.2970$ \\
DiffCom-f & 0.2 & 235 & / & 26.0880 & 0.9266 & 0.1113 & 0.1384 & 47.8137 & $13.7380 \bm{\pm} 0.2188$ \\
DiffCom-s & 0.2 & 50 & / & 25.6508 & 0.9060 & 0.2091 & 0.2047 & 103.4732
& $4.0028 \bm{\pm} 0.0751$ \\
\midrule
SecDiff & 0.2 & 50 & 0.3 & \textbf{26.7672} & \textbf{0.9371} & 0.1273 & 0.1416 & 41.7900 & $3.4004 \bm{\pm} 0.0251$\\
\midrule
\midrule
\multirow{3}{*}{\shortstack[l]{SecDiff\\(Varying $\rho$)}}
& 0.05 & 50 & 0.3 & 26.8197 & 0.9378 & 0.1266 & 0.1414 & 41.7664 & / \\
& 0.1 & 50 & 0.3 & 26.7863 & 0.9373 & 0.1269 & 0.1417 & 41.7856 & / \\
& 0.3 & 50 & 0.3 & 26.7495 & 0.9370 & 0.1274 & 0.1414 & 41.7544 & / \\
\midrule
\multirow{3}{*}{\shortstack[l]{SecDiff\\(Varying $n_{Step}$)}}
& 0.2 & 30 & 0.3 & 26.7726 & 0.9338 & 0.1678 & 0.1665 & 54.7875 & $2.0489 \bm{\pm} 0.0209$\\
& 0.2 & 70 & 0.3 & 26.3940 & 0.9323 & 0.1183 & 0.1402 & 41.6557 & $4.7091 \bm{\pm} 0.0414$ \\
& 0.2 & 100 & 0.3 & 25.7215 & 0.9217 & 0.1232 & 0.1500 & 51.1233 & $6.7542 \bm{\pm} 0.0769$ \\
\midrule
\multirow{3}{*}{\shortstack[l]{SecDiff\\(Varying $w$)}}
& 0.2 & 50 & 0.0 & 26.3493 & 0.9326 & 0.1303 & 0.1435 & 42.9290 & / \\
& 0.2 & 50 & 0.7 & 26.9245 & 0.9391 & 0.1260 & 0.1408 & 41.5939 & / \\
& 0.2 & 50 & 1.0 & 26.9448 & 0.9395 & 0.1255 & 0.1406 & 41.6603 & / \\
\bottomrule
\end{tabular}
\label{tab:hyper-para}
\end{table*}

\subsubsection{Qualitative Comparison}

To further complement the quantitative evaluation, Fig.~\ref{fig:ex1} presents a visual comparison of reconstructed images under the same jamming setting, where $\text{J/R} = 8~\mathrm{dB}$, $\rho = 0.2$, $\mathrm{SNR} = 10~\mathrm{dB}$. The figure includes a clean reference, the degraded jammed version, and the outputs produced by DeepJSCC, three DiffCom variants, and the proposed SecDiff.

The baseline DeepJSCC exhibits severe degradation when subjected to subcarrier jamming. It fails to recover key semantic components, frequently resulting in distorted and textureless faces. In contrast, the DiffCom series shows varying performance depending on its configuration. The full-step DiffCom-n achieves the lowest LPIPS and DISTS values and generates sharp textures. 
However, this apparent sharpness is partly misleading, as the model amplifies residual noise into unnatural high-frequency details, resulting in inconsistent reconstructions. For example, in Fig.~\ref{fig:ex1}, the facial region shows ripple-like noisy patches that are absent in both the original image and the reconstruction produced by SecDiff.
The accelerated DiffCom-f reduces inference time but suffers from perceptual degradation. It fails to completely remove residual noise, producing over-smoothed outputs that deviate noticeably from the original image.
DiffCom-s, configured with the same number of steps as SecDiff, offers comparable runtime but generates images that appear insufficiently denoised, with residual blurry patches similar to those observed in DeepJSCC, highlighting the limitations of simply reducing the diffusion step count.

By comparison, the proposed SecDiff demonstrates superior robustness and visual quality. Even under heavy subcarrier jamming, SecDiff accurately reconstructs fine features, while preserving the overall structure and identity of the image. 

\subsubsection{Parameters Comparison}

Table~\ref{tab:hyper-para} further examines the impact of three key hyperparameters on the performance of SecDiff, the jamming ratio $\rho$, the number of denoising steps $n_{Step}$, and the mask guidance weight $w$. 

First, varying $\rho$ from 0.05 to 0.3 yields only marginal fluctuations in all IQA metrics, with PSNR remaining around $26.8~\mathrm{dB}$ and LPIPS around 0.127. This stability indicates that SecDiff maintains consistent robustness even as the attack intensity increases, confirming the effectiveness of its hybrid guidance under diverse jamming scenarios.

Second, $n_{Step}$ directly influences both the diffusion process and the inference latency. When $n_{Step}$ is reduced to 30, the perceptual quality slightly degrades, as evidenced by an increased LPIPS of 0.1678 and DISTS of 0.1665. Increasing $n_{Step}$ to 70 yields the best trade-off, achieving a low LPIPS of 0.1183 and a competitive FID of 41.66. However, further increasing $n_{Step}$ to 100 results in a PSNR drop to $25.72~\mathrm{dB}$ and an FID increase to 51.12, indicating that excessive iterations may introduce reconstruction artifacts and unnecessarily prolong inference. Considering the runtime of approximately 3~s for $n_{Step}=50$, this configuration provides an optimal balance between reconstruction performance and computational efficiency.

Finally, adjusting $w$, which controls the mask guidance strength, reveals its critical role in enhancing robustness. When $w$ is set to 0, the PSNR drops to $26.35~\mathrm{dB}$ and the FID increases to $42.93$, indicating incomplete recovery of jammed subcarriers. As $w$ increases, the guidance becomes stronger, and values around 0.7 to 1.0 consistently yield the highest MS-SSIM of approximately 0.939 and the lowest FID of approximately 41.6, confirming that strong mask guidance effectively suppresses jamming artifacts without over-constraining the diffusion process.

Overall, the analysis demonstrates that SecDiff is resilient to varying attack ratios, benefits from a moderate number of denoising steps to achieve a favorable trade-off between quality and latency, and requires sufficiently strong mask guidance to fully exploit spectral context for robust image reconstruction.

\begin{table*}[t]\scriptsize
\centering
\caption{
Performance comparison under spoofing attacks with varying parameters at $\text{SNR} = 10~\mathrm{dB}$.
}
\begin{tabular}{lccccccccccc}
\toprule
\textbf{Method} & $n_{step}$ & $n_{m}$ & $\eta$ & Success ratio & PSNR$\uparrow$ & MS-SSIM$\uparrow$ & LPIPS$\downarrow$ & DISTS$\downarrow$ & FID$\downarrow$ & Time (s) \\
\midrule
DeepJSCC & / & / & / & 0\% & 10.6063 & 0.1595 & 0.7599 & 0.4429 & 343.1236 & $0.2773 \bm{\pm} 0.0040$ \\
\midrule
DiffCom-b & 1000 & / & / & 37.33\% & 22.8727 & 0.8796 & 0.1596 & 0.1550 & 75.2779 & $47.9238 \bm{\pm} 1.1594$\\
\midrule
SecDiff & 50 & 5 & 100$\to$1 & \textbf{74.00\%} & \textbf{27.2835} & \textbf{0.9401} & \textbf{0.1201} & \textbf{0.1356}  & \textbf{48.0699} & $6.0892 \bm{\pm} 0.1160$ \\
\midrule
\midrule
\multirow{2}{*}{\shortstack[l]{SecDiff\\(Varying $n_m$)}} 
& 50 & 3 & 100$\to$1 & 3.00\% & 22.6978 & 0.8499 & 0.2631 & 0.2233 & 50.0461 & $4.9600 \bm{\pm} 0.1111$ \\
& 50 & 10 & 100$\to$1 & 66.00\% & 27.2879 & 0.9426 & 0.1227 & 0.1378 & 52.8822 & $10.0980 \bm{\pm} 0.4012$ \\
\midrule
\multirow{2}{*}{\shortstack[l]{SecDiff\\(Varying $\eta$)}} 
& 50 & 5 & 100& 62.00\% & 26.4134 & 0.9341 & 0.1374 & 0.1459 & 59.0733 & / \\
& 50 & 5 & 1 & 54.67\% & 27.4962 & 0.9410 & 0.1237 & 0.1397 & 60.0458 & / \\
\bottomrule
\end{tabular}
\label{tab:blind_diffcom}
\end{table*}

\begin{figure*}[t]
\centering
\begin{subfigure}{.32\textwidth}
  \centering
  \includegraphics[width=0.90\linewidth]{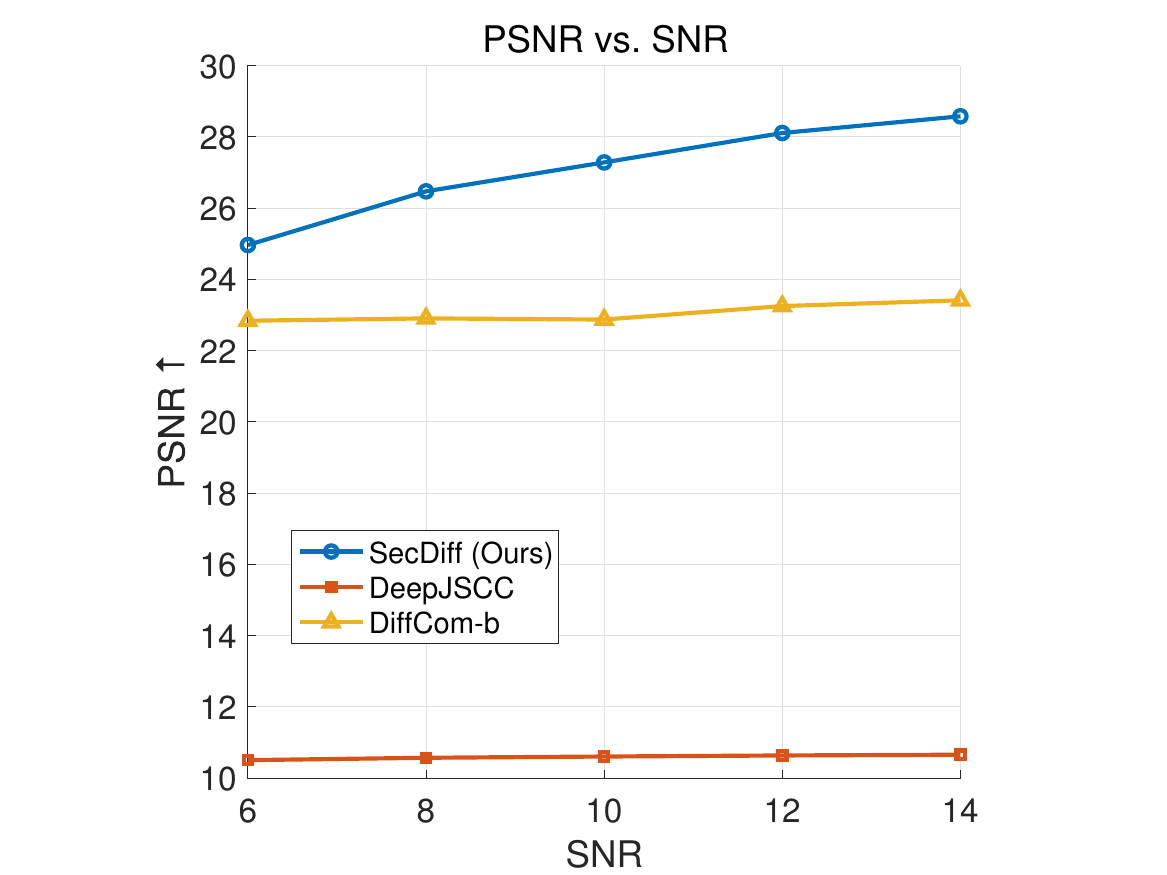}
  \caption{PSNR}
  \label{fig:sub1}
\end{subfigure}
\hfill
\begin{subfigure}{.32\textwidth}
  \centering
  \includegraphics[width=0.90\linewidth]{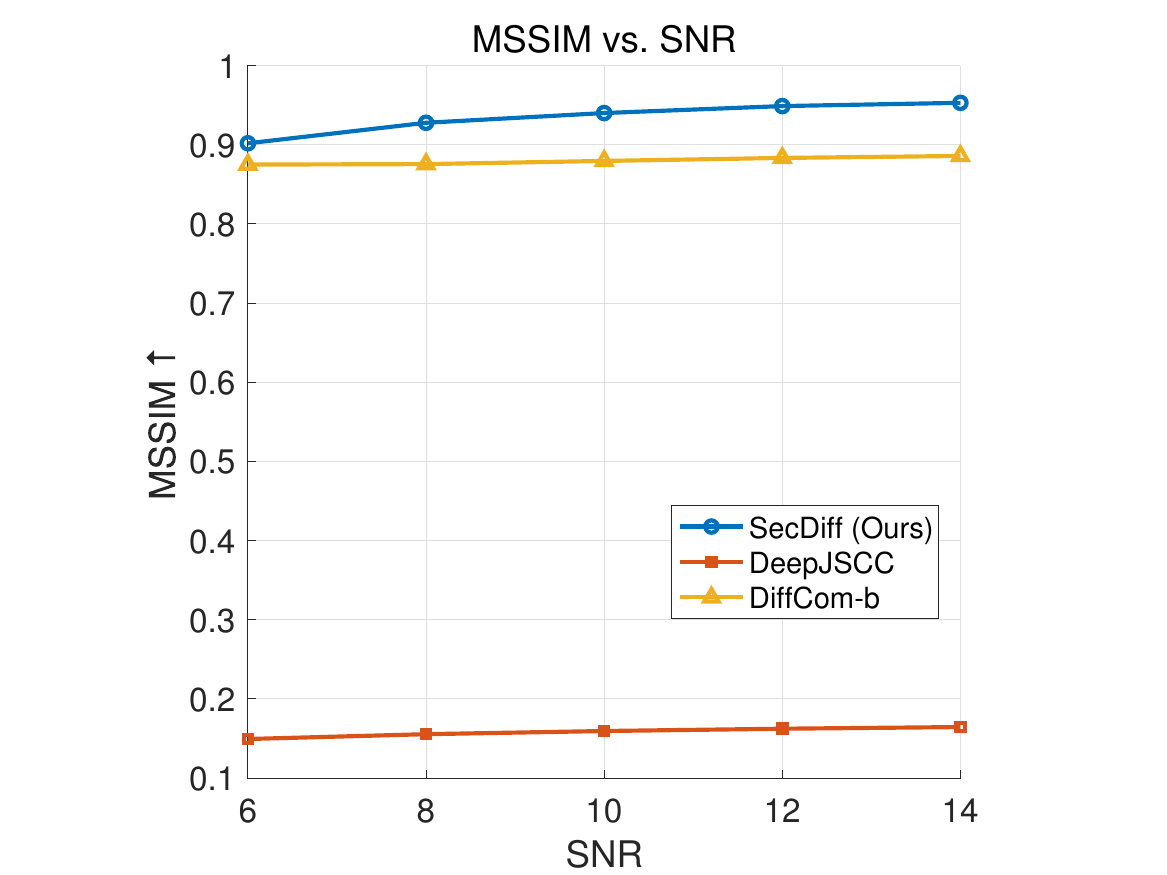} 
  \caption{MS-SSIM}
  \label{fig:sub2}
\end{subfigure}%
\hfill
\begin{subfigure}{.32\textwidth}
  \centering
  \includegraphics[width=0.90\linewidth]{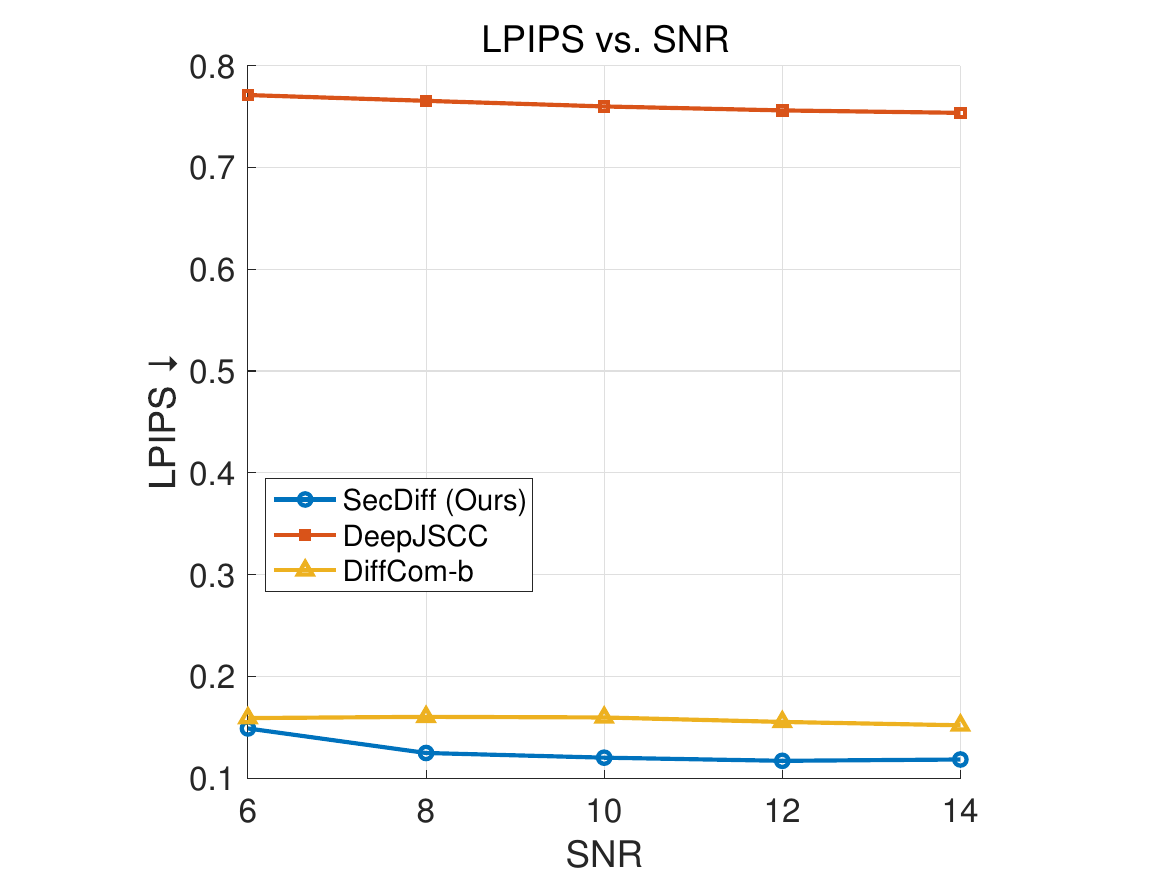}
  \caption{LPIPS}
  \label{fig:sub3}
\end{subfigure}
\vspace{0.4em}
\begin{subfigure}{.32\textwidth}
  \centering
  \includegraphics[width=0.90\linewidth]{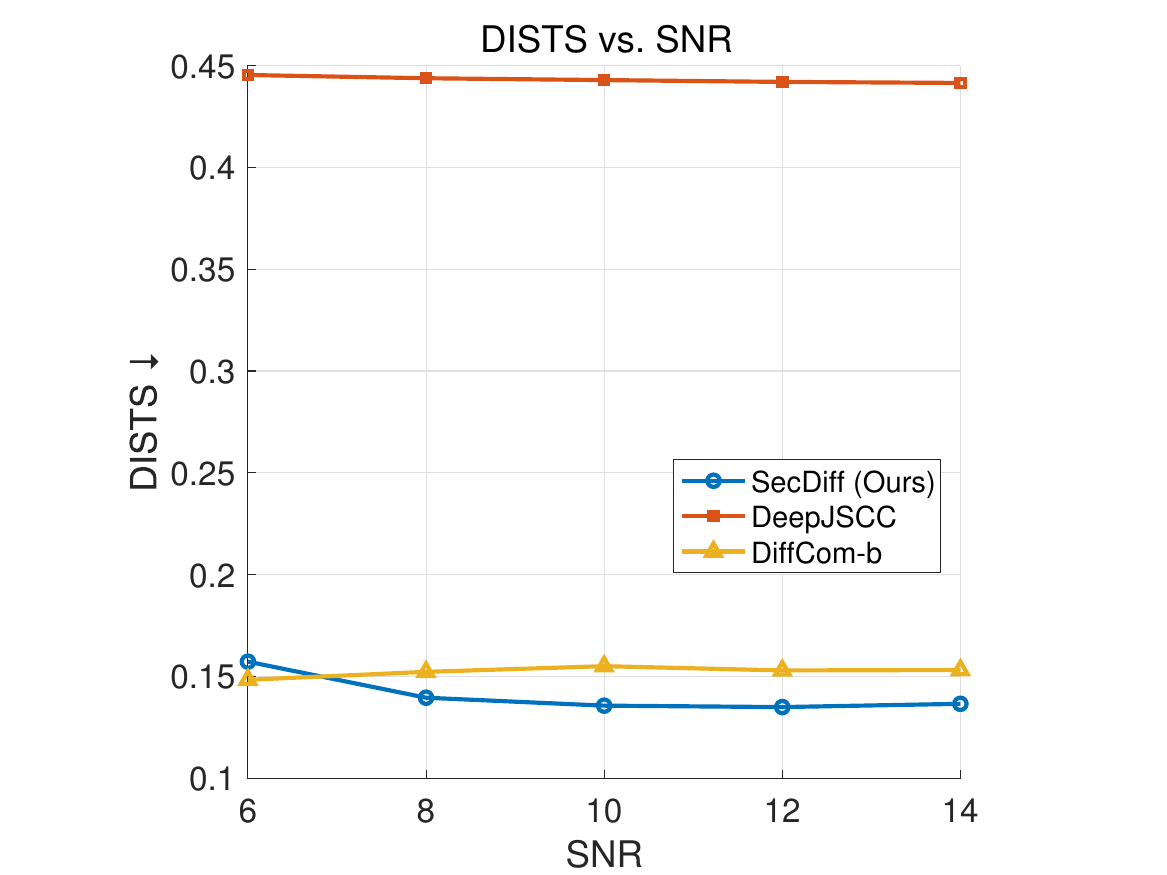}
  \caption{DISTS}
  \label{fig:sub4}
\end{subfigure}
\hfill
\begin{subfigure}{.32\textwidth}
  \centering
  \includegraphics[width=0.90\linewidth]{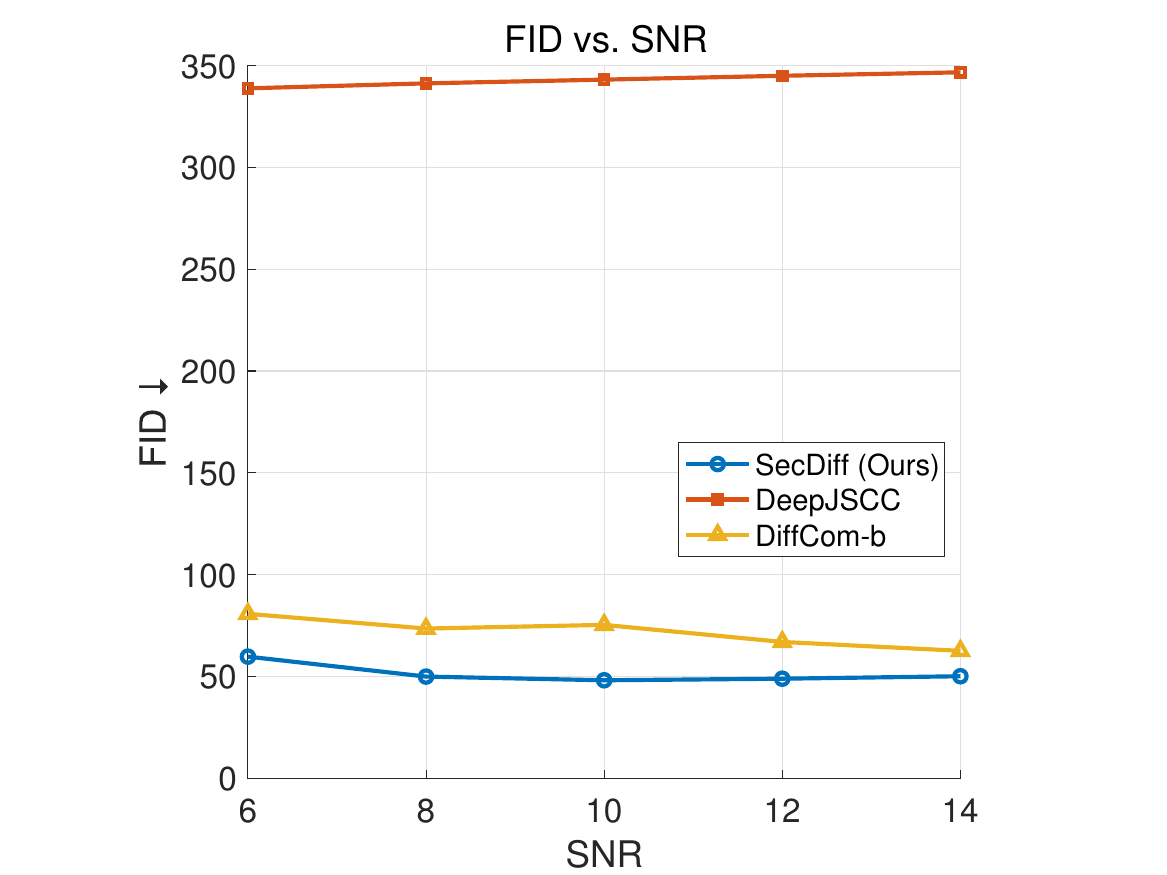}
  \caption{FID}
  \label{fig:sub5}
\end{subfigure}
\begin{subfigure}{.32\textwidth}
  \centering
  \includegraphics[width=0.90\linewidth]{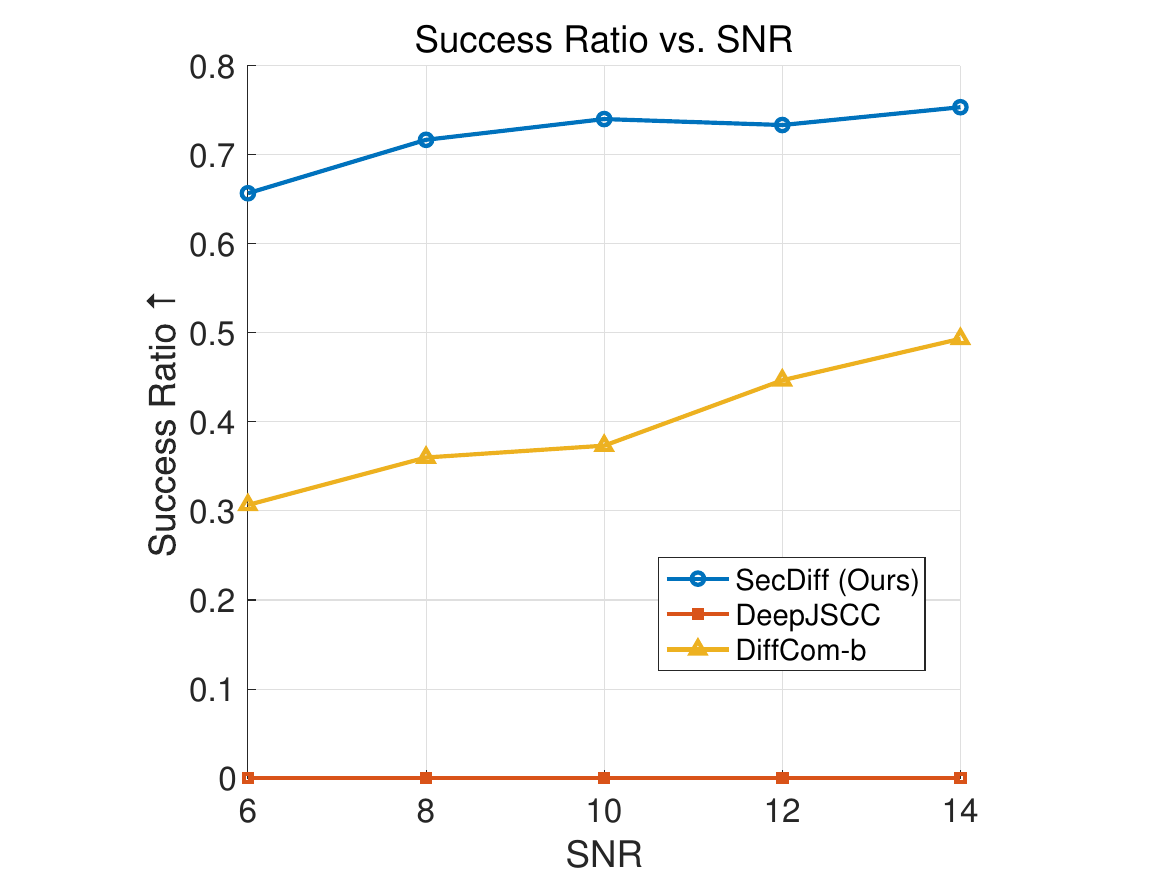}
  \caption{Success ratio}
  \label{fig:sub6}
\end{subfigure}
\caption{Quantitative evaluation of image quality under different $\text{SNR}$ levels with pilot spoofing attacks, using six metrics.}
    \label{fig:six_metrics}
\end{figure*}

\begin{figure*}[htp]
    \centering
    \includegraphics[width= 0.90\linewidth]{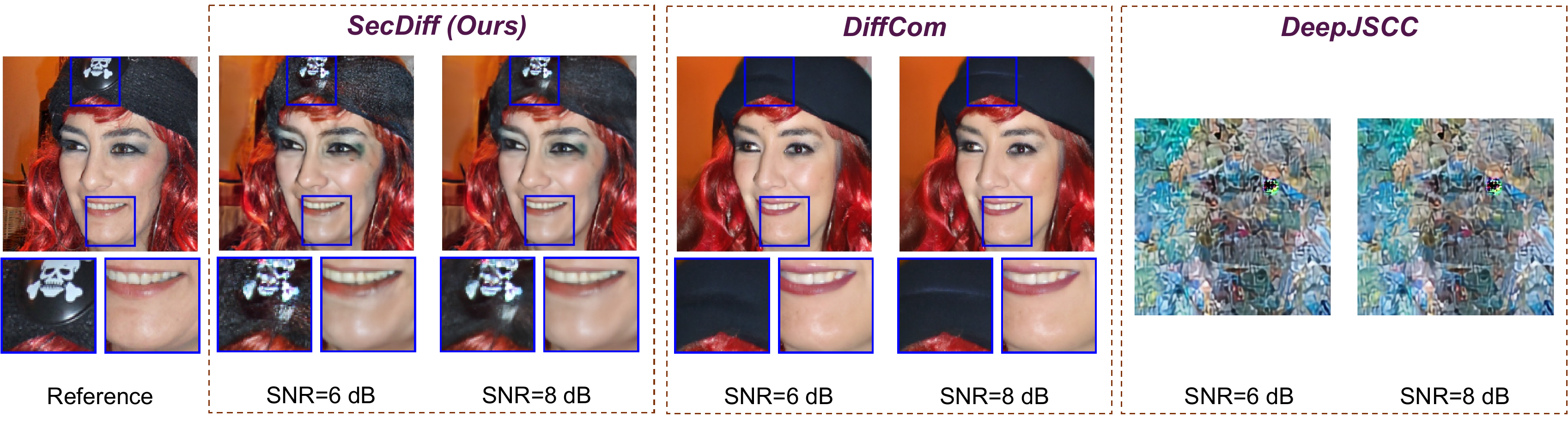}
    \caption{Qualitative comparison of reconstructed images under pilot spoofing attacks.}
    \label{fig:ex2}
\end{figure*}

\vspace{-1mm}

\subsection{Spoofing Defense}


We evaluate the performance of different JSCC frameworks under pilot spoofing attacks. For this experiment, we define a successful recovery as cases where the reconstructed image achieves PSNR $> 20~\mathrm{dB}$. The reported results in Fig.~\ref{fig:six_metrics} correspond to the average performance of SecDiff and DiffCom-b over successfully recovered images.
Additionally, we set the prior regularization weight to $\lambda_2 = 12.0$ and adopt an annealing strategy for the operator learning rate, decreasing $\eta$ from 100 to 1 for SecDiff.

\subsubsection{Quantitative Comparison}

The baseline DeepJSCC completely fails under pilot spoofing, with a 0\% success rate across all SNR levels and a PSNR of only $10.6~\mathrm{dB}$, indicating total semantic collapse and unrecognizable reconstructions.

In contrast, the proposed SecDiff consistently achieves high success rates, rising from 65.7\% at $\mathrm{SNR}=6~\mathrm{dB}$ to 75\% at $\mathrm{SNR}=14~\mathrm{dB}$. 
While DiffCom-b improves over DeepJSCC with a 37.3\% success rate, its performance remains far behind SecDiff, delivering significantly lower perceptual quality.
Compared to DiffCom, SecDiff achieves a 19.3\% higher PSNR, 6.9\% higher MS-SSIM, and reduces LPIPS by 24.7\%, DISTS by 12.5\%, and FID by 36.1\%, at $\mathrm{SNR}=10~\mathrm{dB}$.
These comparisons clearly demonstrate that SecDiff not only achieves substantially higher robustness but also provides consistent gains across all quality metrics compared to both the baseline and DiffCom-b.



Furthermore, DiffCom-b suffers from a long inference time of approximately $47.9$~s per image, which limits its practical applicability. Notably, DiffCom does not support accelerated sampling under blind channel conditions. As a result, even with its full-step configuration, DiffCom-b incurs substantial latency. In comparison, SecDiff completes decoding in only $6.1$~s, providing nearly an eightfold speedup while simultaneously achieving significantly higher success rates and superior performance across all perceptual quality metrics.

\subsubsection{Qualitative Comparison}

Fig. \ref{fig:ex2} presents a visual comparison of reconstructed images at $\mathrm{SNR} = 6~\mathrm{dB}$ and $\mathrm{SNR} = 8~\mathrm{dB}$. The example shown is representative of successful recoveries.

The baseline DeepJSCC fails completely under spoofing attacks, producing outputs that resemble random noise with no discernible semantic content. This collapse is consistent with its 0\% success rate across all SNR levels.
In contrast, DiffCom-b partially restores the global structure of the face but fails to recover fine semantic details. For instance, the hat texture appears oversmoothed, and the facial expression lacks clarity.
By comparison, the proposed SecDiff demonstrates superior robustness and visual quality. SecDiff accurately reconstructs critical features such as the pattern on the hat and the details of the mouth region, while maintaining natural textures and realistic facial expressions. 

\subsubsection{Parameters Comparison}

Table~\ref{tab:blind_diffcom} further investigates the impact of two critical parameters in SecDiff’s EM-guided diffusion framework, the number of EM iterations $n_m$ and the operator learning rate $\eta$. 

First, varying the number of EM iterations $n_m$ shows that this parameter is crucial for successful recovery. When $n_m$ is reduced to 3, the success ratio collapses to 3\%, with PSNR dropping to $22.70~\mathrm{dB}$ and LPIPS increasing to 0.2631, indicating that insufficient channel refinement cannot overcome pilot spoofing. Increasing $n_m$ to 10 improves PSNR to $27.29~\mathrm{dB}$ and maintains a success ratio of 66\%, but at the cost of doubling the runtime to $10.1$~s. The default configuration $n_m=5$ achieves the best balance, delivering the highest success ratio and competitive perceptual quality while keeping runtime moderate.

Second, adjusting the operator learning rate $\eta$, which controls the regularization schedule during channel operator refinement, strongly influences performance. Using a fixed high value $\eta=100$ achieves a success ratio of 62\% but leads to a slightly lower PSNR of $26.41~\mathrm{dB}$ and a higher FID of 59.07 due to over-regularization. Conversely, setting $\eta=1$ results in under-regularization, with only 54.7\% success despite a higher PSNR of $27.50~\mathrm{dB}$. The annealing strategy $\eta=100\to1$ effectively balances stability and adaptiveness, producing the best overall results with the lowest FID of 48.07 and optimal perceptual scores.
\section{Conclusion}

In this work, we have proposed SecDiff, a plug-and-play diffusion-aided decoding framework that enhances the security and robustness of JSCC against adversarial wireless attacks. The framework integrates pseudoinverse-guided sampling with adaptive step-size control, enabling efficient semantic reconstruction under dynamic adversarial conditions.
For subcarrier jamming, we have introduced a power-based masking strategy and recast the recovery task as a masked inpainting problem, where hybrid diffusion guidance leverages spectral context to restore corrupted frequencies. To counter pilot spoofing, we have developed an EM-driven blind channel estimation algorithm that alternates between signal reconstruction and operator refinement, achieving robust decoding without explicit channel state information.
Extensive experiments have demonstrated that SecDiff consistently outperforms existing JSCC and diffusion-based baselines in terms of pixel-level fidelity, perceptual similarity, and semantic realism, while maintaining low inference latency. These results confirm that SecDiff offers a practical and effective solution for secure semantic communication over adversarial OFDM channels.

\bibliography{Ref}

\end{document}